\documentclass{article}

\usepackage[final]{style/neurips_2024}

\usepackage{url}
\usepackage[utf8]{inputenc} %
\usepackage[T1]{fontenc}    %
\usepackage{url}            %
\usepackage{booktabs}       %
\usepackage{enumitem}
\usepackage{amsfonts}       %
\usepackage{nicefrac}       %
\usepackage{microtype}      %
\usepackage{xcolor}         %
\usepackage{enumitem}
\usepackage{xspace}
\usepackage{amsthm}
\usepackage{amsmath,amsfonts,bm}
\usepackage{array}
\makeatletter
\newtheorem*{rep@theorem}{\rep@title}
\newcommand{\newreptheorem}[2]{%
\newenvironment{rep#1}[1]{%
 \def\rep@title{#2 \ref{##1}}%
 \begin{rep@theorem}}%
 {\end{rep@theorem}}}
\makeatother

\newreptheorem{theorem}{Theorem}

\definecolor{myred}{RGB}{215,48,39}
\definecolor{mygreen}{RGB}{26,152,80}
\newcommand{\cmark}{\textcolor{mygreen}{\ding{51}}}
\newcommand{\xmark}{\textcolor{myred}{\ding{55}}}
\newcommand{\halfmark}{\textcolor{gray}{\checkmark\kern-1.1ex\raisebox{.7ex}{\rotatebox[origin=c]{125}{--}}}}
\usepackage[framemethod=TikZ]{mdframed}
\mdfdefinestyle{MyFrame}{%
    linecolor=black,
    outerlinewidth=.3pt,
    roundcorner=5pt,
    innertopmargin=1pt, %
    innerbottommargin=1pt, %
    innerrightmargin=1pt,
    innerleftmargin=1pt,
    backgroundcolor=black!0!white}

\mdfdefinestyle{MyFrame2}{%
    linecolor=white,
    outerlinewidth=1pt,
    roundcorner=1pt,
    innertopmargin=0,
    innerbottommargin=0,
    innerrightmargin=7pt,
    innerleftmargin=7pt,
    backgroundcolor=black!3!white}

\mdfdefinestyle{MyFrameEq}{%
    linecolor=white,
    outerlinewidth=0pt,
    roundcorner=0pt,
    innertopmargin=0pt,
    innerbottommargin=0pt,
    innerrightmargin=7pt,
    innerleftmargin=7pt,
    backgroundcolor=black!3!white}

\newcommand{\RNum}[1]{\uppercase\expandafter{\romannumeral #1\relax}}

\newcommand{\R}{\mathcal{R}}

\newcommand{\vertiii}[1]{{\left\vert\kern-0.25ex\left\vert\kern-0.25ex\left\vert #1 
    \right\vert\kern-0.25ex\right\vert\kern-0.25ex\right\vert}}
\newcommand{\vertiiii}[1]{{\vert\kern-0.25ex\vert\kern-0.25ex\vert #1 
    \vert\kern-0.25ex\vert\kern-0.25ex\vert}}

\usepackage{mathtools}

\newcommand{\xhdr}[1]{{\noindent\bfseries #1.}}
\newcommand{\cut}[1]{}

\newcommand{\angstrom}{\textup{\AA}}

\newcommand{\removelatexerror}{\let\@latex@error\@gobble}

\def\eqref#1{Eq.~\ref{#1}}

\def\1{\bm{1}}

\DeclareMathAlphabet{\mathsfit}{\encodingdefault}{\sfdefault}{m}{sl}
\SetMathAlphabet{\mathsfit}{bold}{\encodingdefault}{\sfdefault}{bx}{n}

\def\gA{{\mathcal{A}}}

\def\gD{{\mathcal{D}}}

\def\gL{{\mathcal{L}}}
\def\gM{{\mathcal{M}}}

\def\gS{{\mathcal{S}}}
\def\gT{{\mathcal{T}}}

\def\sP{{\mathbb{P}}}

\def\R{{\mathbb{R}}}

\newcommand{\sethree}{\mathrm{SE(3)}}

\newcommand{\sothree}{\mathrm{SO(3)}}
\newcommand{\sethreen}{\sethree^{\scriptscriptstyle N}}
\newcommand{\sethreenzero}{\sethree^{\scriptscriptstyle N}_{\scriptscriptstyle 0}}
\newcommand{\sethreezero}{\sethree_{\scriptscriptstyle 0}}

\newcommand{\foldflowpp}[0]{\textsc{FoldFlow-2}\xspace}

\newcommand{\foldflow}[0]{\textsc{FoldFlow}\xspace}

\newcolumntype{C}[1]{>{\centering\arraybackslash}p{#1}}
\newcolumntype{L}[1]{>{\raggedright\arraybackslash}p{#1}}
\newcolumntype{R}[1]{>{\raggedleft\arraybackslash}p{#1}}

\usepackage{todonotes}
\usepackage{amssymb,fge}
\usepackage{thm-restate}
\usepackage{wrapfig}
\usepackage{bbm}
\usepackage{mathrsfs}
\usepackage{soul}
\usepackage{array}
\usepackage{multirow}
\usepackage{algorithm}
\usepackage[commentColor=black,beginLComment=/*~, endLComment=~*/]{algpseudocodex}
\usepackage{subcaption}
\usepackage{pifont}%
\usepackage[final,pagebackref=true]{hyperref} %
\usepackage{cleveref}
\usepackage{float} %
\renewcommand*{\backrefalt}[4]{%
    \ifcase #1 \footnotesize{(Not cited.)}%
    \or        \footnotesize{(Cited on page~#2)}%
    \else      \footnotesize{(Cited on pages~#2)}%
    \fi}
\newcolumntype{P}[1]{>{\centering\arraybackslash}p{#1}}

\hypersetup{
	colorlinks=true,       %
	linkcolor=blue,        %
	citecolor=blue,        %
	filecolor=magenta,     %
	urlcolor=blue         
}

\title{Sequence-Augmented $\sethree$-Flow Matching For Conditional Protein Backbone Generation}

\author{%
Guillaume Huguet$^{1,2,3}$\thanks{Co-first and corresponding authors: \{\texttt{guillaume.huguet,james\}@dreamfold.ai}},
James Vuckovic$^{1}$\footnotemark[1],
Kilian Fatras$^{1}$\thanks{Core contributor},
Eric Thibodeau-Laufer$^{1}$\footnotemark[2], \\
\textbf{
Pablo Lemos$^{1}$,
Riashat Islam$^{1}$,
Cheng-Hao Liu$^{1,3,4}$,
Jarrid Rector-Brooks$^{1,2,3}$,
} \\
\textbf{
Tara Akhound-Sadegh$^{1,2,4}$, Michael Bronstein$^{1,5,6}$, Alexander Tong$^{1,2,3}$\thanks{Equal advising}, Avishek Joey Bose$^{1,5}$\footnotemark[3]
}\\
$^1$Dreamfold, $^2$Université de Montréal,  $^3$Mila,  $^4$McGill University, $^5$University of Oxford, $^6$Aithyra
}

\begin{document}

\maketitle

\begin{abstract}
\looseness=-1
Proteins are essential for almost all biological processes and derive their diverse functions from complex $3 \rm D$ structures, which are in turn determined by their amino acid sequences. 
In this paper, we exploit the rich biological inductive bias of amino acid sequences and introduce \foldflowpp\footnote{Our code can be found at \url{https://github.com/DreamFold/FoldFlow}}, a novel sequence-conditioned $\sethree$-equivariant flow matching model for protein structure generation. \foldflowpp presents substantial new architectural features over the previous \foldflow family of models including a protein large language model to encode sequence, a new multi-modal fusion trunk that combines structure and sequence representations, and a geometric transformer based decoder. To increase 
diversity and novelty of generated samples---crucial for de-novo drug design---we
train \foldflowpp at scale on a new dataset 
that is an order of magnitude 
larger than PDB datasets of prior works, containing both known proteins in PDB and high-quality synthetic structures achieved through filtering. We further demonstrate the ability to align \foldflowpp to arbitrary rewards, e.g. increasing secondary structures diversity, by introducing a Reinforced Finetuning (ReFT) objective. We empirically observe that \foldflowpp outperforms previous state-of-the-art protein structure-based generative models, improving over RFDiffusion in terms of unconditional generation across all metrics including designability, diversity, and novelty across all protein lengths, as well as exhibiting generalization on the task of equilibrium conformation sampling. Finally, we demonstrate that a fine-tuned \foldflowpp makes progress on challenging conditional design tasks such as designing scaffolds for the VHH nanobody.
\end{abstract} 

\vspace{-1em}
\section{Introduction}
\label{sec:introduction}

\looseness=-1
Rational design of novel protein structures via generative modeling holds significant promise for accelerating computational drug discovery~\citep{chevalier2017massively,ebrahimi2023engineering}. In particular, the ability to design proteins with a pre-specified functional property is arguably one of the principal tools in addressing global health challenges such as COVID-19~\citep{cao2020novo,gainza2023novo}, influenza~\citep{strauch2017computational}, and cancer~\citep{silva2019novo}. In many instances, designing function involves the design of both the $3 \rm D$ geometric structure of the protein as well as its specific chemical interactions. In proteins, the amino-acid sequences determine the interaction between protein backbones and side chains, which fold into a distribution of protein structures. Consequently, the functional properties of protein structures can be \emph{inferred} from its sequence.

\looseness=-1
The representation of proteins plays a key aspect in any computational approach to protein engineering. The $3 \rm D$ structure of proteins can be mathematically represented on the space of rotation and translation invariant $\sethreen$. Several unconditional protein generative models have been developed recently to generate new protein backbones \citep{yim2023se, bose2023se3stochastic}. While these models demonstrate the ability to design new proteins, they are insufficiently tailored for downstream drug discovery applications, where the primary challenge lies in generating proteins that are specifically tailored to interact effectively with a given target. 
In real-world drug design problems, one often knows the target protein (its amino acid sequence and often an experimentally verified $3 \rm D$ structure). 
In machine learning terms, the design of new proteins (``{\em de novo}'' design) that can drug the given target can be framed as a \emph{conditional} generation problem. 
This raises the following research question: 
\begin{center} 
{\em How can we leverage the structure and sequence of a target to inform {\em de novo} protein design?}
\end{center}

\looseness=-1
\xhdr{Current work}
In this paper, we introduce \foldflowpp a novel protein structure generative model that is additionally conditioned on protein sequences.
\foldflowpp is built on the foundations of \foldflow \citep{bose2023se3stochastic} and is an $\sethreen$-invariant generative model for protein backbone generation that handles multi-modal data by design. Specifically, \foldflowpp introduces several new architectural components over previous protein structure generative models that
enable it to process both $3\rm D$ structure and discrete sequences. 
These include (1) a joint structure and sequence encoder; (2) a multi-modal fusion trunk that combines the representations from each modality in a shared representation space; and (3) a transformer-based geometric decoder. In contrast to prior efforts to incorporate sequences in structure-based generative models~\citep{campbell2024generative}, \foldflowpp leverages the representational power of a large pre-trained protein language model in ESM~\citep{lin2022language} enabling it to make use of the rich biological inductive bias found in sequences but at a scale far beyond ground-truth experimental $3\rm D$ structures found in the Protein Data Bank (PDB). 

\looseness=-1
As a sequence-conditioned model, \foldflowpp is able to tackle a suite of new tasks beyond simple unconditional generation. Specifically, our model can additionally be used for protein folding by simply generating structures conditioned on sequence as well as hard, biologically motivated conditional design problems. For instance, 
our model can perform partial structure generation by conditioning on a masked sequence, i.e., structure in-painting. This enables \foldflowpp to be better equipped than prior structure-only generative models to tackle the key challenges in de novo drug design. For example, in settings where we aim to engineer a structure that binds and neutralizes a desired target protein structure and sequence pair; this is precisely a structure and sequence in-painting problem.

\looseness=-1
As diversity and quantity of training samples play a crucial role in downstream generative modeling performance on conditional design tasks, we construct a new large dataset---an order of magnitude larger than PDB---of high-quality synthetic structures filtered from SwissProt~\citep{jumper2021highly,varadi2021alphafold}. We further investigate the impact of fine-tuning \foldflowpp using Reinforced Fine-Tuning (ReFT), a new approach that aligns flow-matching generative models to arbitrary rewards. In the context of protein backbone generation, we apply fine-tuning to improve the properties of generated backbones, such as optimizing for the diversity of secondary structures, as well as improving the performance on conditional generation tasks like generating scaffolds around a target motif.

\xhdr{Main results} We summarize the main empirical results obtained using \foldflowpp below:
\begin{itemize}[noitemsep,topsep=0pt,parsep=0pt,partopsep=0pt,label={\large\textbullet},leftmargin=*]

\item \looseness=-1 We empirically demonstrate that \foldflowpp achieves state-of-the art performance for unconditional generation and leads to the most \emph{designable, novel, and diverse} proteins. In particular, \foldflowpp improves over RFDiffusion~\citep{watson_novo_2023} and FoldFlow~\citep{bose2023se3stochastic}.
\item \looseness=-1 We find \foldflowpp closes the gap in performance with purpose-built folding models like ESMFold and improves by a factor of $\approx 4 \times$ compared to MultiFlow~\citep{campbell2024generative}, the most comparable protein structure generative model that also leverages sequences.
\item We use \foldflowpp to solve a biologically relevant conditional design problem in motif scaffolding. We find that a fine-tuned \foldflowpp is able to solve all $24/24$ scaffolds in the benchmark dataset from \citet{watson_novo_2023}. On challenging VHH nanobodies, it solves $9/25$ refoldable motifs in comparison to $5/25$ for the previous best approach RFDiffusion.
\item \looseness=-1 We hypothesize that \foldflowpp is able to perform zero-shot equilibrium conformation sampling on unseen proteins in the ATLAS molecular dynamics (MD) dataset~\citep{vander2024atlas} based on conformation variation seen within the protein data bank. We observe that \foldflowpp is able to capture different modes of the equilibrium conformation comparably to ESMFlow-MD~\citep{jing2024alphafold}, a model fine-tuned on MD data, but lags behind AlphaFlow-MD~\citep{jing2024alphafold}.

\end{itemize}

\section{Background and preliminaries}
\label{sec:background}
\looseness=-1

\subsection{Protein backbone and sequence parametrizations}
\label{sec:protein_backbone_parametrization}
\looseness=-1
\xhdr{Sequence representation}
Protein sequences correspond to the chain of amino acids,  which for a protein of length $N$ is identified by a discrete token $a^i \in \{ 1, \dots,  20\} =: \gA$. As is customary in protein language models~\citep{lin2022language}, these discrete tokens are encoded using a one-hot representation. We denote the entire amino acid sequence associated with a protein as $\text{A} \in \R^{N \times 20}$.

\looseness=-1
\xhdr{Structure representation} The $3\rm D$ structure of protein backbones can be represented as rigid frames associated with each residue in an amino acid sequence~\citep{jumper2021highly}. Each residue, $i$, within a protein backbone of length $N$ consists of idealized coordinates of their $4$ heavy atoms $\text{N}^*,\text{C}^*_{\alpha}, \text{C}^*,\text{O}^* \in \R^3$, with $\text{C}^*_{\alpha} = (0,0,0)$. The defining property of rigid frames is that they can be viewed as elements of the special Euclidean group $\sethree$ and as such each frame $x = (r, s) \in \sethree$ contains a rotation $r$ and translation $s$ component. Applying a rigid transformation $x^i$ to the idealized coordinates of the heavy atoms allows us to represent the rigid frame of a given residue, $[\text{N},\text{C}_{\alpha},\text{C},\text{O}]^i = x^i \circ [\text{N}^*,\text{C}^*_{\alpha},\text{C}^*,\text{O}^*]$, where $\circ$ is the binary operator associated to the group, which for $\sethree$ is simply matrix multiplication. This leads to a structure representation of the complete $3\rm D$ coordinates associated with all heavy atoms of a protein as the tensor $\text{X} \in \R^{N \times 4 \times 3}$.

\looseness=-1
\xhdr{$\sethree$: the group of rigid motions} The special Euclidean group $\sethree$ contains rotations and translations in three dimensions and can be thought of in several ways. It is a Lie group, i.e., a differentiable manifold endowed with a group structure. $\sethree$ can be seen as the group of rigid frames, representing 3D rotations and translations. As a Lie group, $\sethree$ can be uniquely identified with its Lie algebra, the tangent space at the identity element of the group. $\sethree$ is also a matrix Lie group, meaning that its elements can be represented with matrices. It can formally be written as the semidirect product of the rotation and the translation groups, $\sethree \cong \sothree \ltimes (\R^3, +)$. A more detailed introduction to Riemannian manifolds and Lie theory, with an emphasis on $\sethree$ is provided in~\S\ref{app:riemannian_geo_primer}.

\subsection{Flow matching on the $\sethree$ group}
\vspace{-5pt}
\looseness=-1
As Lie groups are smooth manifolds, they can also be equipped with a Riemannian metric, which can be used to define distances and geodesics on the manifold. On $\sethree$, a natural choice of the metric decomposes into the metrics on its constituent subgroups, $\sothree$ and $\R^3$~\citep{bose2023se3stochastic, yim2023se}. This allows us to build independent flows on the group of rotations and translations and induce a flow directly on $\sethree$. As flow matching on Euclidean spaces is well-studied~\citep{albergo_building_2023,lipman_flow_2022,liu_flow_2023}, we restrict our focus on reviewing flows, conditional probability paths, and vector fields over the group $\sothree$.  

\cut{With a judicious choice of Riemannian metric, it is possible to decompose $\sethree$ into its constituent rotation and translation subgroups $\sethree \cong \sothree \ltimes (\R^3, +)$. Loosely speaking this results as a consequence of the disintegration of measures for $\sethree$-invariant measures~\citep{pollard2002user}. Such a fact allows us to perform flow-matching on the group of rotations $\sothree$ and translations $\R^3$ independently and induce a flow directly on $\sethree$.}

\cut{
Given the parametrization of the backbone as $N$ rigid frames, our goal is to model an $\sethree^N$ invariant density using a flow-matching objective.  The translation-invariance is achieved by constructing the flow on the subspace $\sethree_0^N$, where the center of mass of the inputs is removed. Additionally, as on the product manifold, consisting of $N$ copies of $\sethree_0$ the metric extends naturally, we can focus on building flows on the group of rotations $\sothree$ and translations $\R^3$, for each of the residues independently. As flow matching on Euclidean spaces is well-studied~\citep{albergo_building_2023,lipman_flow_2022,tong2023improving} we restrict our focus on reviewing flows, conditional probability paths, and vector fields over the group of rotations $\sothree$.  
}

\looseness=-1
\xhdr{Probability paths on $\sothree$}
Given two densities $\rho_0, \rho_1 \in \sothree$, a probability path $\rho_t: [0, 1] \to \sP(\sothree)$ is an interpolation, parametrized by time, $t$, between the two densities in probability space. Without loss of generality, we may consider $\rho_0$ to be the target data distribution and $\rho_1$ an easy-to-sample source distribution. A \emph{flow} is a one-parameter diffeomorphism in $t$, $\psi_t: \sothree \to \sothree$. It is the solution to the ordinary differential equation (ODE): $ \frac{d}{dt} \psi_t(r) = u_t \left(\psi_t(r) \right)$, with initial condition $\psi_0(r) = r$, where $u_t$ is the time-dependent smooth vector $u_t: [0,1]\times\sothree \to \sothree$.  It is said that $\psi_t$ {\em generates} $\rho_t$ if it induces a pushforward map $\rho_t=[\psi_t]_\#(\rho_0)$.

\cut{
push-forward map $\psi_t: \sothree \to \sothree$---i.e., a one parameter diffeomorphism in $t$---satisfies the identity $\rho_t=[\psi_t]_\#(\rho_0)$ it is called a \emph{flow} that generates $\rho_t$. Additionally, $\psi_t$ is the solution ordinary differential equation (ODE): $ \frac{d}{dt} \psi_t(r) = u_t \left(\psi_t(r) \right)$, with initial conditions $\psi_0(r) = r$, and the time-dependent smooth vector $u_t: [0,1]\times\sothree \to \sothree$. 
}

\looseness=-1
\xhdr{Matching vector fields on $\sothree$}
The framework of Riemannian flow matching~\citep{chen2024riemannian} can also accommodate Lie groups such as $\sothree$. Consequently, to learn a continuous normalizing flow (CNF) that pushes forward samples $r_0 \sim \rho_0$ to $ r_1 \sim \rho_1$ we must regress a parametric vector field $v_{\theta} \in  \mathfrak{X}(\sothree)$ in the tangent space of the manifold to the target conditional vector field $u_t(r_t | r_0, r_1)$, for all $t \in [0,1]$. Conveniently, the target $u_t(r_t | r_0, r_1)$ is the time derivative of a point $r_t$ along the shortest path between $r_0$ and $r_1$---i.e., the geodesic interpolant $r_t = \exp_{r_0}(t \log_{r_0}(r_1))$. Furthermore, for $\sothree$ the target conditional vector field admits a closed-form expression $u_t(r_t | r_0, r_1) = \log_{r_t}(r_0) / t$ as the exponential and logarithmic maps are numerically computable using the axis-angle representation of the group elements~\citep{bose2023se3stochastic}. Given these ingredients, we can formulate the flow matching objective for $\sothree$ as:
\begin{equation}\label{eq:foldflowbase}
\gL_{\sothree}(\theta)=\mathbb{E}_{t\sim \mathcal{U}(0,1), q(r_0,r_1), \rho_t(r_t | r_0, r_1)} \left \|v_\theta(t, r_t) - \log_{r_t}(r_0)/t\right\|_{\sothree}^2. 
\end{equation}
\looseness=-1
In \cref{eq:foldflowbase}, $q(r_0, r_1)$ is any coupling between samples from the source and target distributions. An optimal choice is to set $q(r_0, r_1) = \pi(r_0, r_1)$ which is the coupling, $\pi$, that solves the Riemannian optimal transport problem using minibatches~\citep{bose2023se3stochastic, tong2023improving, fatras20a}.
Finally, the generation of samples is done by first drawing from a source sample $r_1 \sim \rho_1$ and integrating the ODE backward in time using the learned vectorfield $v_{\theta}$.

\section{\foldflowpp}%
\begin{wraptable}{r}{0.5\textwidth}
    \vspace{-15pt}
    \centering
    \caption{\looseness=-1 \small Overview of the conditioning capability of unconditional ($\emptyset$), folding ($\rm A$), and inpainting ($\rm A, \rm X$) of various protein backbone generation models.}
\resizebox{0.5\textwidth}{!}{
\begin{tabular}{l|cccc}
\toprule
Method  & $\emptyset$ & $\rm A$ & $(\rm A, \rm X)$ \\
\midrule
AlphaFold~\citep{jumper2021highly}           & \xmark & \cmark & \xmark\\
RFDiffusion~\citep{watson_novo_2023}         & \cmark & \xmark & \xmark\\
Chroma~\citep{ingraham2023illuminating}      & \cmark & \xmark & \xmark\\
FrameDiff~\citep{yim2023se}                  & \cmark & \xmark & \xmark\\
\foldflow~\citep{bose2023se3stochastic}      & \cmark & \xmark & \xmark\\
FrameFlow~\citep{yim2023fast}                & \cmark & \xmark & \cmark \\

Motif RFDiffusion~\citep{watson_novo_2023}   & \xmark & \xmark & \cmark \\
Multiflow~\citep{campbell2024generative}     & \cmark & \cmark & \cmark\\
\foldflowpp (Ours)                           & \cmark & \cmark & \cmark \\
\bottomrule
\end{tabular}
}
    \label{tab:conditioning}
\end{wraptable}
\label{sec:ffpp}
\looseness=-1 We now present \foldflowpp, our sequence-conditioned structure generative model. 
\foldflowpp operates on protein backbones $x_0 \sim \rho_0$ which are parametrized as $N$ rigid frames as
well as their corresponding sequence $a$. As protein backbones contain symmetries from $\sethree$, we design \foldflowpp as an $\sethreen$-invariant density using a flow-matching objective. We achieve translation invariance by constructing the flow on the subspace $\sethreenzero$, where the center of mass of the inputs is removed. Additionally, we can focus on building flows on the group of rotations $\sothree$ and translations $\R^3$, for each of the $N$ residues independently, as $\sethreenzero$ can be viewed as a product manifold consisting of $N$ copies of $\sethreezero$. The overall loss function for the model decomposes into per residue rotation and translation losses $\gL = \gL_{\sothree} + \gL_{\R^3}$,
\begin{equation}\label{eq:foldflowpploss}
\gL = \mathbb{E}_{t\sim \mathcal{U}(0,1), \rho_t(x_t | x_0, x_1, \bar{a})} \left[\left \|v_\theta(t, r_t, \bar{a}) - \log_{r_t}\frac{r_0}{t}\right\|_{\sothree}^2  + \left \|v_\theta(t, s_t, \bar{a}) - \frac{(s_t - s_0)}{t}\right\|_{2}^2\right],
\end{equation}
\looseness=-1
where the pair $(x_0, x_1) \sim \pi(x_0, x_1)$ is sampled from the optimal transport plan $\pi$. In addition, the sequence $\bar{a} = a \odot m$ corresponds to $x_0$ and is masked completely, with a mask $m$, with a probability $\rm{Bern}(0.5)$. Operationally, this means $50\%$ of the time the model is trained unconditionally with no sequence information, i.e., $\bar{a} = [\emptyset]^N$, while the other $50\%$ the model has access to the full sequence $\bar{a} = a$. Optimizing the loss in \cref{eq:foldflowpploss} is equivalent to maximizing the conditional log-likelihood of observing protein structures given their sequences $\log p(\rm X | \rm A)$ when the sequence is not masked and maximizing the unconditional log-likelihood $\log p(\rm X)$ when the sequence is fully masked. Due to the ability to mask sequences, \foldflowpp enables new modeling capabilities in comparison to existing models as outlined in \cref{tab:conditioning}. More precisely, \foldflowpp trained using masked sequences can perform a diverse set of tasks, outlined in \cref{tab:ffpp_capabilities}, beyond simple unconditional backbone generation which aids in tackling more biologically relevant problems that require conditional generation such as mimicking a protein folding model and designing the $3\rm D$ scaffolds around a target motif. 

\begin{wraptable}{r}{0.5\textwidth}
    \centering
    \vspace{-10pt}
    \caption{\small By manipulating the input modalities, \foldflowpp is able to perform a diverse set of conditional and unconditional generation tasks including biologically relevant tasks such as designing scaffolds.}
    \vspace{-5pt}
    \resizebox{0.5\textwidth}{!}{
    \begin{tabular}{l l l l l}
        \toprule
        & \textbf{Task Name} & \textbf{Sequence Inputs} & \textbf{Structure Inputs} \\
        \midrule
        (T1) & Unconditional & Fully-masked & Noise \\
        (T2) & Folding & Unmasked & Noise \\
        (T3) & In-Painting & Partially Masked & Partially Masked\\
        \bottomrule
    \end{tabular}
    }
    \vspace{-10pt}
    \label{tab:ffpp_capabilities}
\end{wraptable}

\looseness=-1
With the breadth of tasks $\textbf{T1--T3}$ (table~\ref{tab:ffpp_capabilities}) \foldflowpp unlocks new structural design capabilities beyond the simple unconditional generation ability of \foldflow.
We next outline the architectural components of \foldflowpp in~\S\ref{sec:architecture} before detailing the training procedure in~\S\ref{sec:training}, which includes key design decisions regarding the construction of our new scaled dataset of ground truth PDBs and filtered AlphaFold2 synthetic structures. We also outline the inference procedure for sampling in~\S\ref{app:inference_details}. We conclude by discussing various techniques to fine-tune \foldflowpp, including methods based on filtering with auxiliary rewards for supervised fine-tuning~\S\ref{sec:finetuning} to align protein structures.

\begin{figure}
    \centering
    \begin{tabular}{c}
        \includegraphics[width=1.0\textwidth]{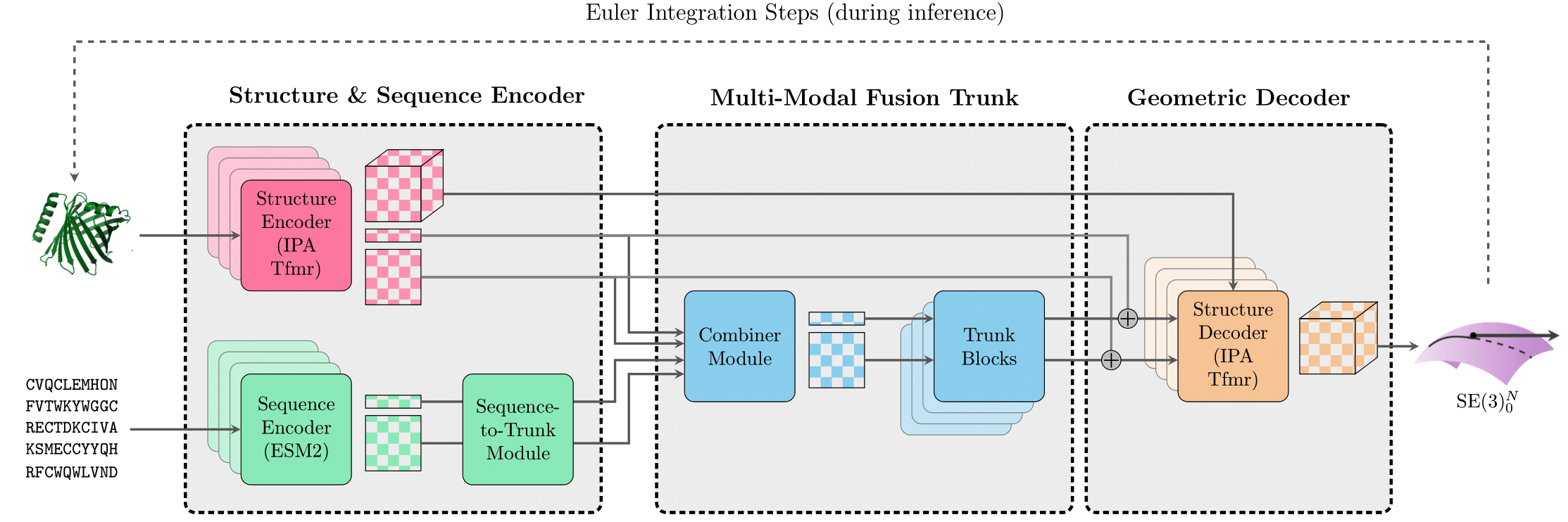}\\
        \begin{tabular}{c c c}
            \includegraphics[width=3.75cm]{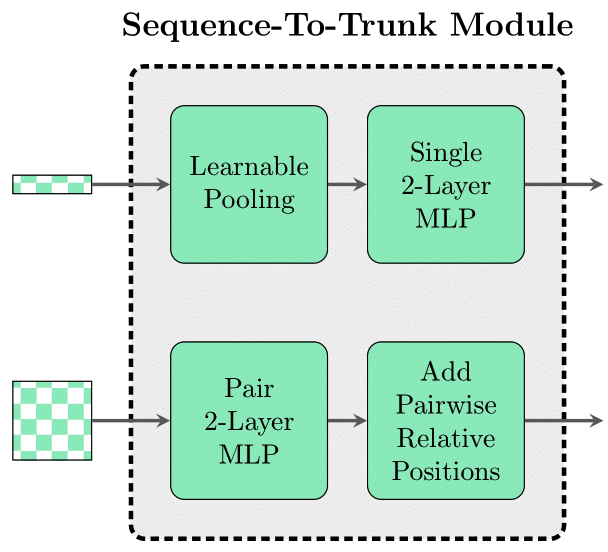} &
            \includegraphics[width=4.2cm]{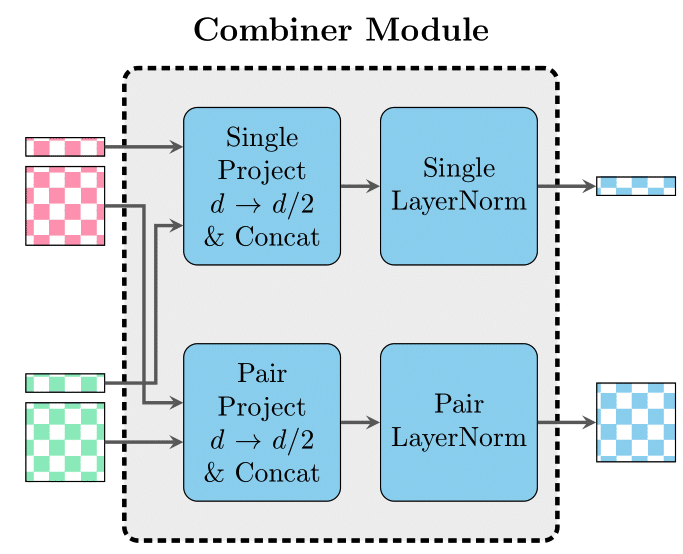} &
            \raisebox{0.5cm}{\includegraphics[width=2.5cm]{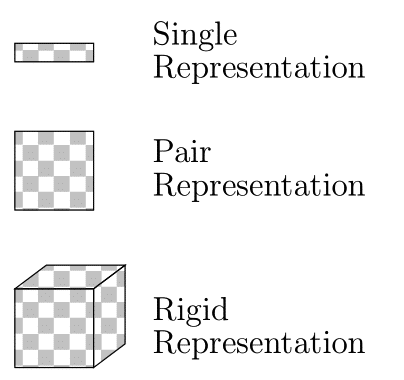}}
        \end{tabular}
    \end{tabular}
    \caption{\small \foldflowpp architecture which processes sequence and structure and outputs $\sethreenzero$ vectorfields.}
        \label{fig:schematic}
    \vspace{-10pt}
\end{figure}

\subsection{\foldflowpp Architecture}
\label{sec:architecture}
\looseness=-1
The \foldflowpp architecture is comprised of three core components: (1) \textbf{Structure and sequence encoder:} An encoder which encodes both structures and sequences; (2) \textbf{Multi-modal fusion trunk:} the trunk which combines the multi-modal representations of the encoded structure and sequences; and (3) \textbf{Geometric Decoder:} a decoder that consumes the fused representation from the trunk and outputs a generated structure. The overall architecture of \foldflowpp is depicted in \cref{fig:schematic}. %

\looseness=-1
\xhdr{Structure and Sequence Encoder}
We leverage existing state-of-the-art architectures to encode the structure and sequence modalities separately. For structure encoding, we rely on the invariant point attention (IPA) transformer architecture~\citep{jumper2021highly}, which is $\sethree$-equivariant. The benefit of the IPA architecture is that it is highly flexible and can both consume and produce a structure---i.e., $N$ rigid frames---\emph{and} also output single and pair representations of the input structure. 

\looseness=-1
To encode amino-acid sequences, we use a large pre-trained protein language model: the $650\rm M$ variant of the ESM2 sequence model~\citep{lin2022language}. Large protein language models have a strong inductive bias on atomic-level predictions of protein structures while exhibiting strong generalization properties beyond any known experimental structures---which we argue is highly correlated with goals of \emph{de novo} structure design. Moreover, the ESM2 architecture also produces single and pair representations for an encoded sequence of amino acids, which conceptually correspond to the single and pair representations from the structure encoder. Consequently, the output space of each modality prescribes a natural fusion of representations into a joint single and pair latent space for a given input protein.

\looseness=-1
\xhdr{Multi-Modal fusion trunk} 
After encoding both input structure and sequence, we construct a joint representation for the single and pair representation using a ``project and concatenate'' combiner module with simple MLPs, see \cref{fig:schematic}. We use LayerNorm~\citep{ba2016layer} throughout the architecture as it is essential to accommodate differently-scaled inputs. The joint representations are further processed by a series of Folding blocks~\citep{lin2023evolutionary}, which refines the single and pair representations via triangular self-attention updates. %

\xhdr{Geometric decoder} 
To decode the joint representations of the inputs into $\sethreenzero$ vector fields, we once again leverage the IPA Transformer architecture. The decoder takes as input the single, pair outputs of the trunk \emph{and} the rigid representations from the structure encoder. One of our major findings is that including a skip-connection between the structure encoder and the decoder is essential for good performance as the temporal information is only given to the structure encoder. 

Given each component, we stack $2-2-2$ blocks for the encoder, trunk, and decoder components.

\subsection{Training}
\label{sec:training}

We train \foldflowpp by alternating between both folding and unconditional generation tasks using a novel sequence-and-structure flow matching procedure, described below.

\looseness=-1
\xhdr{Dataset construction}
The generalization ability of generative models trained using maximum likelihood is determined by the quality and diversity of curated training data~\citep{kadkhodaie2023generalization}. Due to the limited size of ground truth structures in the Protein Data Bank (PDB) we aim to improve training set diversity by additionally curating a dataset of filtered AlphaFold2 structures from SwissProt~\citep{jumper2021highly,varadi2021alphafold}. To ensure \foldflowpp is trained on high-quality synthetic structures, we employ a set of stringent filtering techniques that remove many undesignable proteins from SwissProt. After filtering, our final dataset consists of $160K$ structures and constitutes approximately an 8$\times$ fold increase compared to prior works~\citep{yim2023se,bose2023se3stochastic}. Our exact layered filtering strategy for synthetic structures in SwissProt is outlined by the following steps:

\begin{enumerate}[label=\textbf{(Step \arabic*)},left=0pt,nosep]
    \item  \looseness=-1 \xhdr{Filtering low-confidence structures} We use per-residue local confidence metrics like the average pLDDT to filter out low-confidence structures from the initial SwissProt dataset. %
    \item \looseness=-1\xhdr{Masking low-confidence residues} Globally high-confident structures may include low-confidence residues with disordered regions that can impede training. We use a per-residue pLDDT threshold of $70$ to mask such ``low-quality" residues during training.     
    \item \looseness=-1 \xhdr{Filter high-confidence, low-quality structures} The nature of synthetic data means that even following steps 1 and 2 low-quality data persists in a curated dataset. To combat this we further filter structures by learning a light-weight structure prediction model trained on structural features predictive of protein quality.   
\end{enumerate}
\looseness=-1
We report a detailed analysis of each step in the filtration process in~\S\ref{app:dataset_filtering} which includes examples of low-quality structures that were filtered as illustrative examples. The impact of these findings is empirically corroborated in by analyzing generated samples from \foldflowpp in~\S\ref{sec:ablation}.

\looseness=-1
During training, we set the fraction of synthetic samples that may be seen during an epoch to $2/3$ of the epoch. This prevents the model from overfitting to the remaining noise in the synthetic data, and is also common practice when training with synthetic data~\citep{hsu2022learning,lin2023evolutionary}. Anecdotally, we did not notice an improvement from using a smaller proportion of synthetic structures. Finally, in the \foldflowpp architecture, we keep the ESM pre-trained language model fixed during training and train all other components (encoder, trunk, and decoder) from scratch. The results presented in table \ref{tab:main_table} and \S\ref{sec:motif_scaffolding} use PDB data only, as this displayed the best performance for designability scores.

\subsection{Fine-Tuning \foldflowpp}
\label{sec:finetuning}

\looseness=-1
We explore the efficacy of fine-tuning \foldflowpp with preferential alignment. 
We take a supervised fine-tuning approach \citep{WeiBZGYLDDL22} that uses an additional fine-tuning dataset which is filtered using pre-specified auxiliary rewards $r_{\rm aux}$ to create a preferential dataset $\gD_{\rm pref}$. We term this Reinforced FineTuning (ReFT) since fine-tuning in this manner can be considered aligning \foldflowpp generations to the auxiliary reward. Summarizing this in three steps: (1) We take a curated dataset of proteins with desirable metrics; (2) We use $r_{\rm aux}$ to score the samples from step 1 and filter them to get a subset of high-scoring samples; (3) We then improve \foldflowpp by SFT on the filtered subset. Finetuning with ReFT optimizes the following optimization objective $ \mathcal{L}_{\textsc{ReFT}}(\theta)$, 
\begin{equation}
    \max_{p_{\theta}} \mathcal{L}_{\textsc{ReFT}}(\theta)  =  \mathbb{E}_{(x,a) \sim \gD_{\rm pref}} \left[ r_{\rm aux}(x)  \log p_{\theta}(x | a) \right] .
\end{equation}
\looseness=-1
Compared to recent alignment methods based on reward models, as in RLHF \citep{BaiRLHF}, ReFT uses a filtered structure dataset to fine-tune \foldflowpp. Standard RL approaches seek to fine-tune generative model-based model-generated data and assume access to evaluating the reward function. Our approach maximizing $ \mathcal{L}_{\textsc{ReFT}}(\theta)$ requires constructing $\gD_{\rm pref}$ with auxiliary reward $r_{\rm aux}$, demonstrated by the improvement in secondary structure diversity in \S\ref{sec:increasing_secondary_structure_diversity}.

\section{Experiments}\label{sec:experiments}

\looseness=-1
We evaluate \foldflowpp on multiple protein design tasks including unconditional generation, motif scaffolding, folding, fine-tuning to improve secondary structure diversity, and equilibrium conformation sampling from molecular dynamics trajectories. We provide implementation details in~\S\ref{app:experimental_details}.

\looseness=-1
\xhdr{Baselines} As our main baselines for the unconditional generation task we rely on pre-trained versions of FrameDiff~\citep{yim2023se}, Chroma~\citep{ingraham2023illuminating}, Genie~\citep{lin2023generating}, MultiFlow~\citep{campbell2024generative}, and RFDiffusion which is the current gold standard~\citep{watson_novo_2023}. 
In conditional generation tasks like motif scaffolding, we compare against a conditional variant of FrameFlow~\citep{yim2023fast} as well as RFDiffusion.
For protein folding, we focus on comparing against ESMFold~\citep{lin2022language} and MultiFlow which also leverages sequence information. Lastly, for conformational sampling the principal baselines are ESMFlow and AlphaFlow~\citep{jing2024alphafold}.

\subsection{Unconditional protein backbone generation}
\label{sec:exp_unconditional}
\looseness=-1

\looseness=-1
We evaluate unconditional structure generation using metrics that assess the designability, novelty, and diversity of generated structures. For each method, we generate $50$ proteins at lengths $\{100, 150, 200, 250, 300\}$ (c.f.\ \foldflowpp samples in \cref{fig:generated_samples_ff}). Designability is computed by using the \emph{self-consistency} metric which compares the refolded proteins (with ProteinMPNN~\citep{dauparas_robust_2022} and ESMFold~\citep{lin2022language}) with the original one. Novelty is computed using: 1.) the fraction of designable proteins with TM-score $<0.3$ and 2.) the average maximum TM-score of designable generated proteins to the training data. Finally, diversity uses the average pairwise TM-score designable samples averaged across lengths as well as the maximum number of clusters.

\begin{figure}
    \centering
    \includegraphics[width=1.0\textwidth]{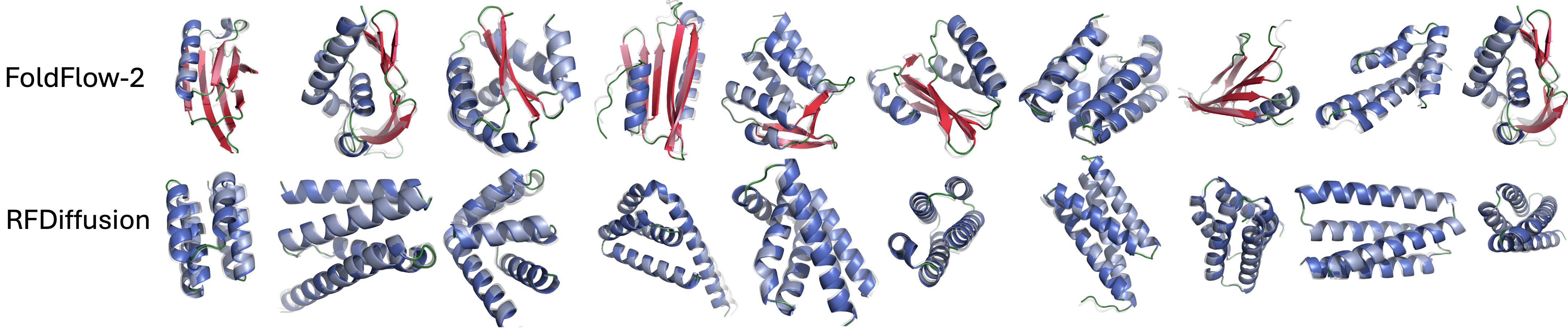}
    \caption{\small Uncurated designable (scRMSD $< 2\angstrom$) length 100 structures with ESMFold refolded structure from \foldflowpp and RFDiffusion colored by secondary structure assignment. \foldflowpp is significantly more diverse in terms of secondary structure composition where we see RFDiffusion generates mostly $\alpha$-helices.}
    \label{fig:samples}
    \vspace{-15pt}
\end{figure}

\begin{table}[htb]
    \centering
    \vspace{-2pt}
    \caption{\looseness=-1\small Comparison of Designability (fraction with scRMSD < $2.0\angstrom$), Novelty (max.\ TM-score to PDB and fraction of proteins with averaged max.\ TM-score $<0.3$ and scRMSD $<2.0\angstrom$), and Diversity (avg.\ pairwise TM-score and MaxCluster fraction). Designability and Novelty metrics include standard errors.%
    }
    \resizebox{1\textwidth}{!}{
    \begin{tabular}{lrrrrrrr}
        \toprule
          \multicolumn{1}{c}{} & \multicolumn{1}{c}{Designability}  & \multicolumn{2}{c}{Novelty} & \multicolumn{2}{c}{Diversity} \\
          \cmidrule(lr){2-2} \cmidrule(lr){3-4} \cmidrule(lr){5-6}
        & Frac.\ $<2\angstrom$ ($\uparrow$)  & Frac.\ TM $< 0.3$ ($\uparrow$) & avg.\ max TM ($\downarrow$) & pairwise TM ($\downarrow$)& MaxCluster ($\uparrow$)\\
        \midrule
        RFDiffusion             & 0.969 $\pm$ 0.023 & 0.116 $\pm$ 0.020 & 0.449 $\pm$ 0.012 & 0.256 & 0.172 \\
        Chroma                  & 0.636 $\pm$ 0.030 & 0.214 $\pm$ 0.033 & 0.412 $\pm$ 0.011 & 0.272 & 0.132 \\
        Genie                   & 0.581 $\pm$ 0.064 & 0.120 $\pm$ 0.021 & 0.434 $\pm$ 0.016 & 0.228 & 0.274 \\
        FrameDiff              & 0.402 $\pm$ 0.062  & 0.020 $\pm$ 0.009 & 0.542 $\pm$ 0.046 & 0.237 & 0.310\\ %
        \foldflow              & 0.820 $\pm$ 0.037  & 0.188 $\pm$ 0.025 & 0.460 $\pm$ 0.020 & 0.247 & 0.228\\ %
        
        \foldflowpp & 0.976 $\pm$ 0.010  & 0.368 $\pm$ 0.031 & 0.363 $\pm$ 0.009 & 0.205 & 0.348\\ %
        \bottomrule
        \end{tabular}
        }

    \label{tab:main_table}
\end{table}

\looseness=-1
\xhdr{Results} We see that \foldflowpp outperforms all other methods on all metrics---crucially without requiring a pretrained folding model as part of the architecture like RFDiffusion. In particular, we observe that \foldflowpp produces the most designable samples with $97.6\%$ of samples being refolded by ESMFold to within $<2\angstrom$. We also find that \foldflowpp novelty improves over RFDiffusion by an absolute $25.2 \%$ in the fraction of designable samples with TM-score $<0.3$. Furthermore, we observe $19.9\%$ and $102.3\%$ relative improvement in the diversity of \foldflowpp over RFDiffusion as measured by the pairwise TM-score and Max Cluster fraction. This places \foldflowpp as the current \emph{most designable, novel, and diverse} protein structure generative model. 

\begin{figure}[ht]
    \centering
    \begin{tabular}{c c c}
\begin{subfigure}[b]{0.3\textwidth}
        \centering
        \includegraphics[width=1\textwidth]{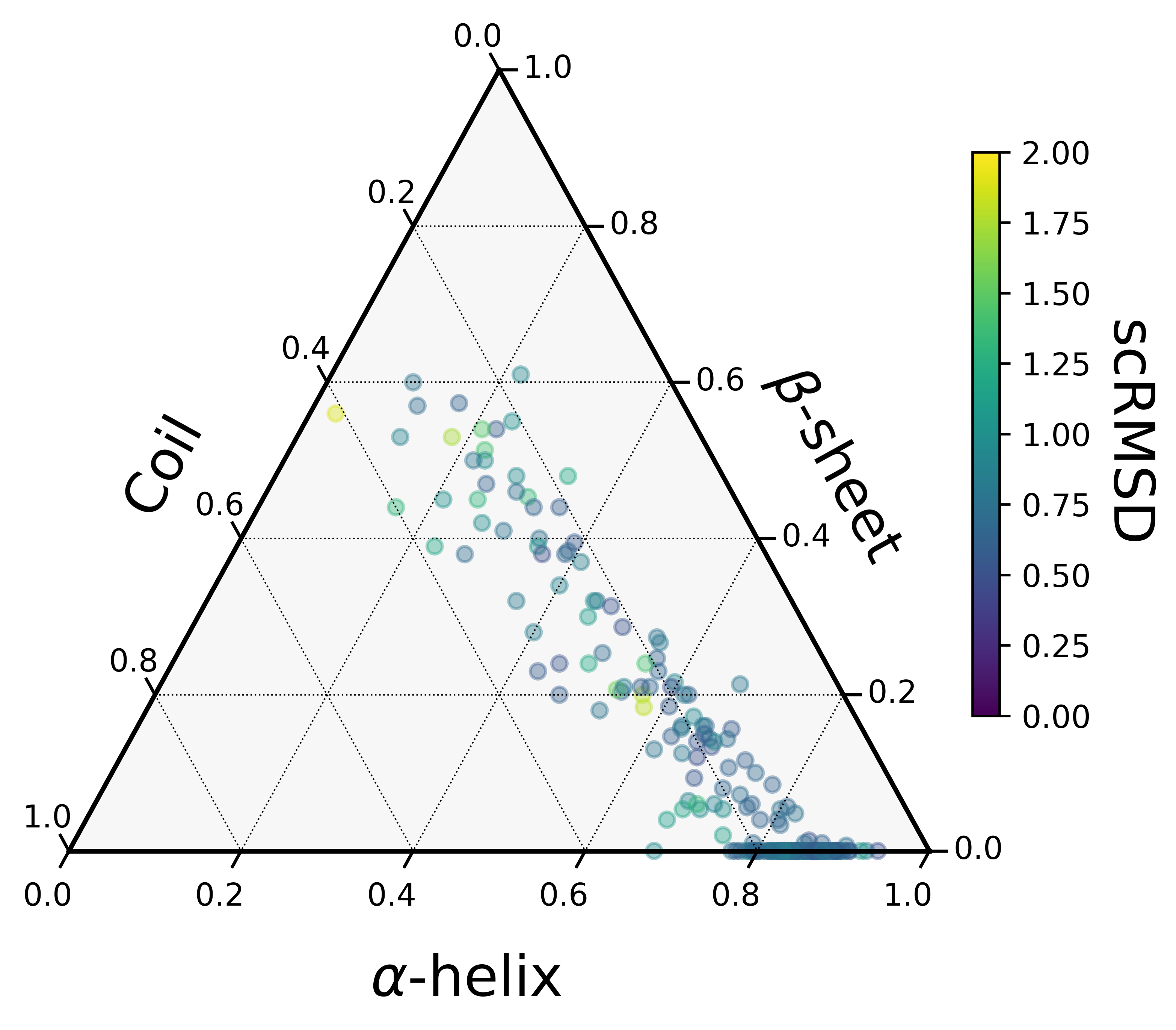}
        \caption{\foldflowpp}
    \end{subfigure}
    \begin{subfigure}[b]{0.3\textwidth}
        \centering
        \includegraphics[width=1\textwidth]{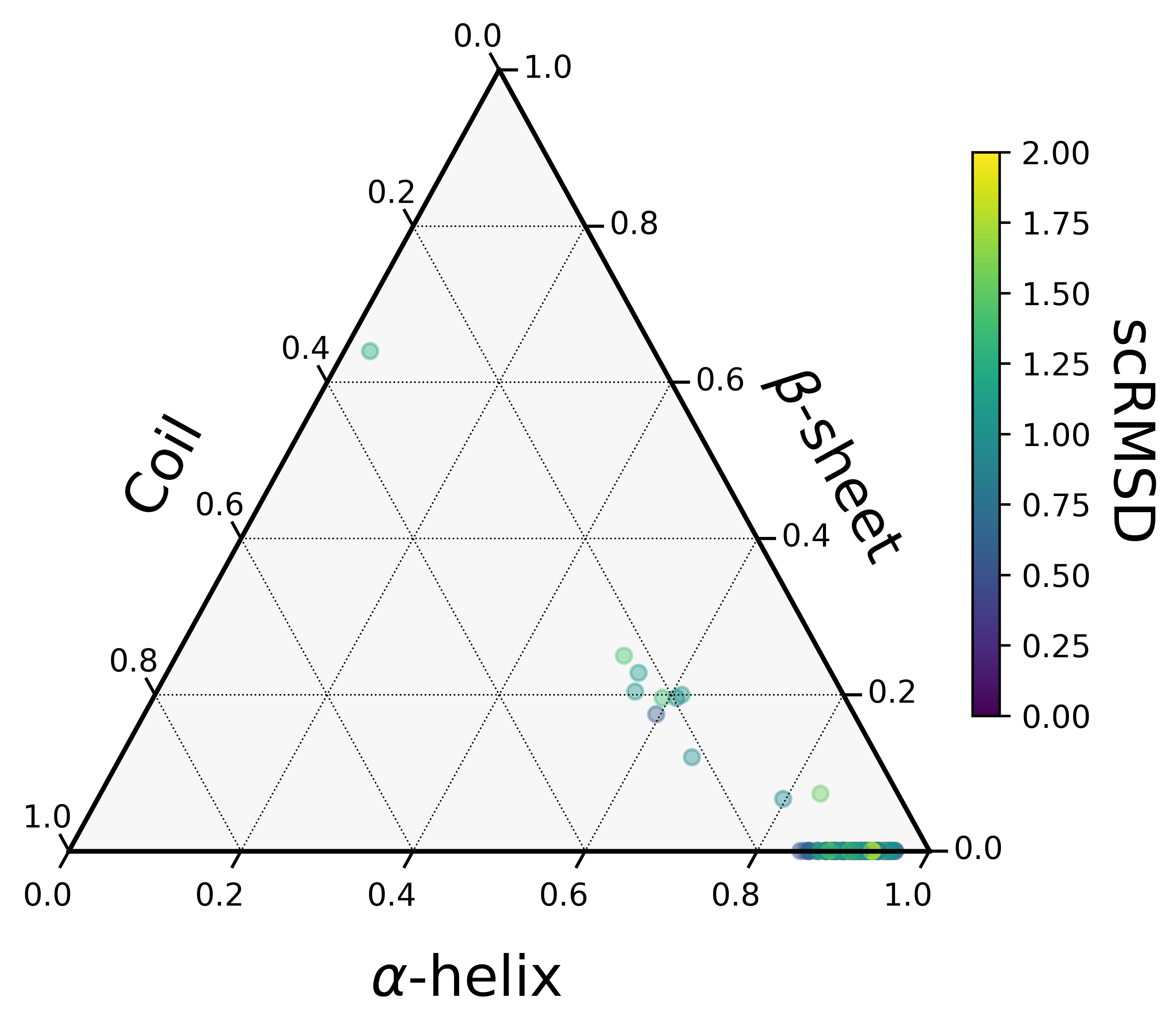}
        \caption{\foldflow}
    \end{subfigure} 
    \begin{subfigure}[b]{0.3\textwidth}
        \centering
        \includegraphics[width=1\textwidth]{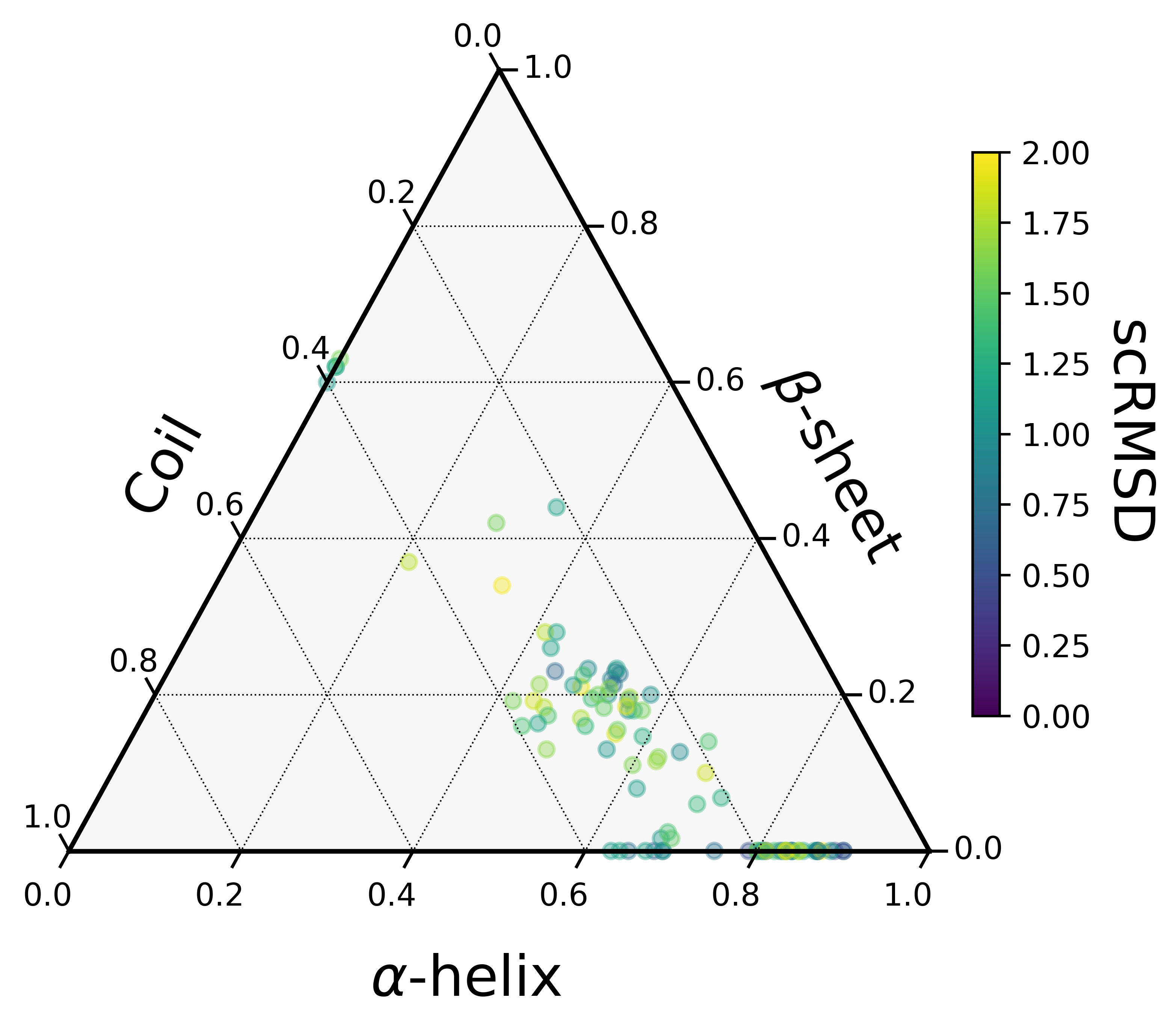}
        \caption{FrameDiff}
    \end{subfigure} \\
    \begin{subfigure}[b]{0.3\textwidth}
        \centering
        \includegraphics[width=1\textwidth]{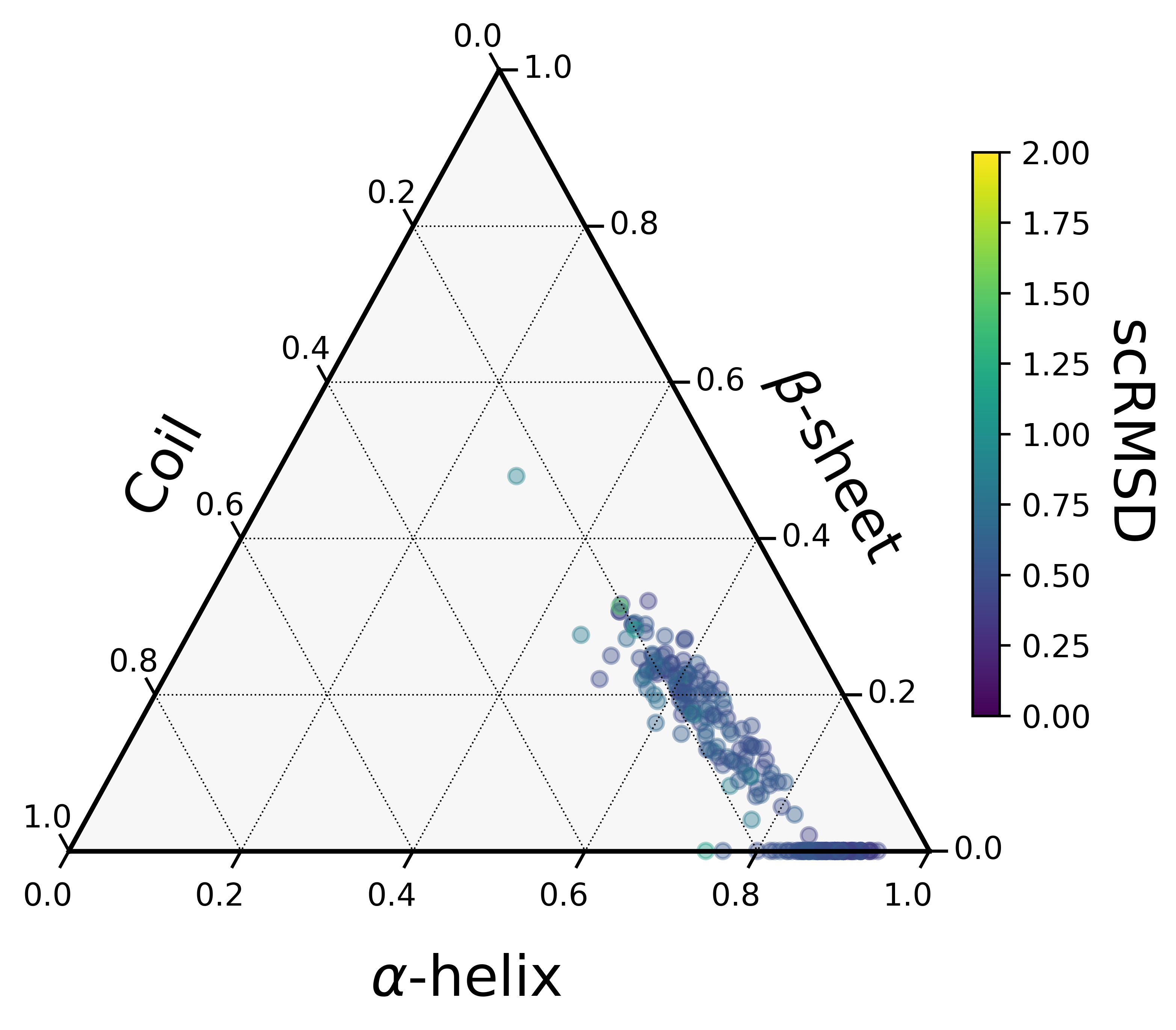}
        \caption{RFDiffusion}
    \end{subfigure}
       \begin{subfigure}[b]{0.3\textwidth}
        \centering
        \includegraphics[width=1\textwidth]{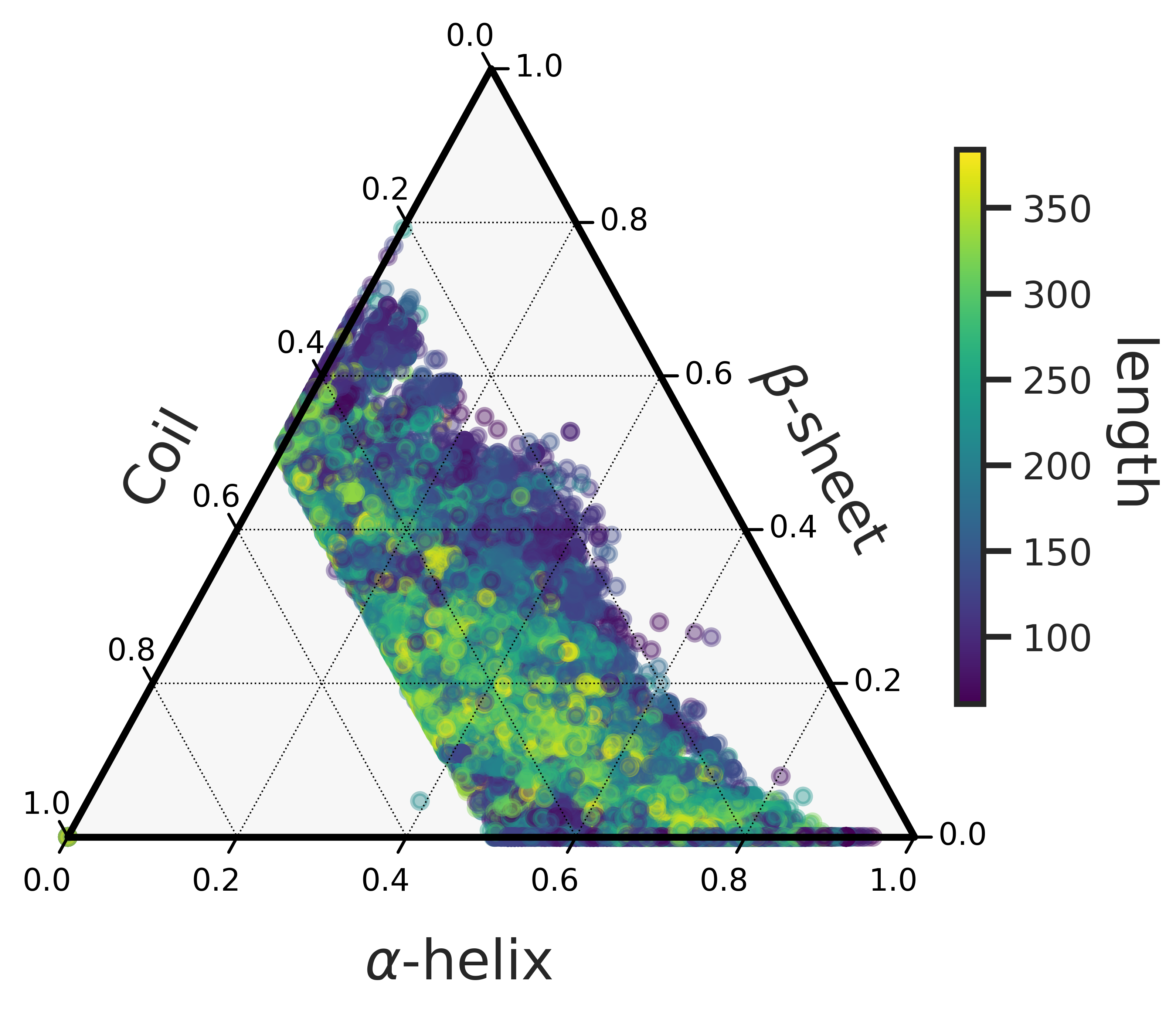}
        \caption{Data}
    \end{subfigure}
       \begin{subfigure}[b]{0.3\textwidth}
        \centering
        \includegraphics[width=1\textwidth]{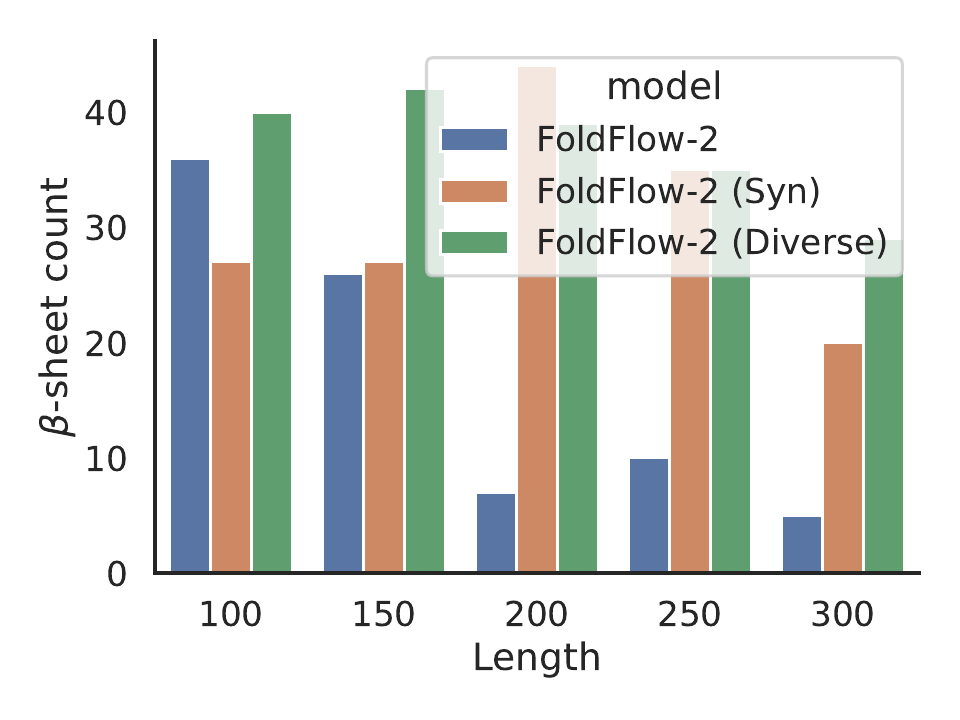}
        \caption{$\beta$-sheet Count}
    \end{subfigure}
    \end{tabular}
    \caption{\small Distribution of secondary structure elements ($\alpha$-helices, $\beta$-sheets, and coils) of designable (scRMSD $< 2.0$) proteins generated by  various models. \foldflowpp generates more diverse designable backbones.}
        \label{fig:secondary_structure_diversity_triangle}
    \vspace{-5pt}
\end{figure}

\looseness=-1
We present uncurated generated samples of \foldflowpp and RFDiffusion in \cref{fig:samples}. We further visualize the distribution of secondary structures of all methods in \cref{fig:secondary_structure_diversity_triangle}. We see a clear indication that \foldflowpp is able to produce the most diverse secondary structures---more closely matching the training distribution (see \cref{fig:secondary_structure_diversity_triangle}e)---and improving over RFDiffusion. We further observe increased amounts of $\beta$-sheets and coils which are particularly challenging for models like FrameDiff and \foldflow that primarily generate $\alpha$-helices. 
We also include multiple ablations on architectural choices, inference annealing, and sequence conditioning in \cref{tab:ablation}.

\subsection{Increasing secondary structure diversity with finetuning}
\label{sec:increasing_secondary_structure_diversity}
\looseness=-1
We investigate ReFT based data filtering to improve diversity of secondary structures in generated samples. We use a diversity score based auxiliary reward for filtering, based on weighted entropy on the proportions of each residue belonging to each type of secondary structure---i.e., alpha-helices ($\alpha$), coils $\rm c$, beta-sheets $\beta$ in the set $\gS$, that can be analytically written as $ r_{\rm diversity} = \left(\sum_{s \in \gS} p_s w_s\right ) \left(1 + \sum_{s \in \gS} p_s \log p_s \right)$. 
Due to models producing increasing amounts of helices, we use $w_{\alpha}=1$, $w_c=0.5$ and $w_{\beta} = 2$, and take top $25\%$ of samples according to the $r_{\rm diversity}$.  
Experimental results in \cref{fig:secondary_structure_diversity_triangle}f with generated samples in \cref{fig:generated_samples_ff} demonstrate that protein at all lengths benefit from training with ReFT as measured by diversity of generated samples, and produces most amount of $\beta$-sheets, and can surpass diversity improvement already obtained by training using synthetic structures as in \cref{fig:secondary_structure_diversity_triangle}.

\subsection{Folding sequences}
\label{sec:folding}
\begin{wraptable}{r}{0.35\textwidth}
    \caption{\small Folding model evaluation on a test set of 268 proteins from PDB.}
    \vspace{-5pt}
    \label{tab:my_label}
    \centering
    \resizebox{0.35\textwidth}{!}{
    \begin{tabular}{lr}
    \toprule
    Model &  RMSD ($\downarrow$) \\
    \midrule
        ESMFold & 2.322 $\pm$ 4.270 \\
        MultiFlow & 14.995 $\pm$ 3.977 \\
        \foldflowpp & 3.237 $\pm$ 4.145 \\
    \bottomrule
    \end{tabular}}
    \label{tab:folding_main}
\end{wraptable}

\looseness=-1
Given that \foldflowpp is sequence conditioned, we can perform protein folding by providing a valid sequence during inference. During training, \foldflowpp tries to transform a $\sethreenzero$ noise sample into the given sequence's 3D structure. Therefore, we aim to measure the generalization properties of our model to fold unseen sequences. We evaluate folding on a test set of $268$ unseen proteins from the PDB dataset. We compare the folding capabilities of \foldflowpp, ESMFold, and Multiflow. In \cref{tab:folding_main}, we report the aligned RMSD between the predicted backbone and the ground truth backbone. We find that \foldflowpp, trained for structure generation, approaches the performance of ESMFold which is a purpose-built folding model. We contextualize this result by noting that \foldflowpp $\approx 4 \times$ is better at folding than the most comparable model in MultiFlow~\citep{campbell2024generative} which is a multi-modal flow matching model using sequences.%

\subsection{Motif Scaffolding}
\label{sec:motif_scaffolding}
\looseness=-1
In motif scaffolding, we are tasked with designing a subset of residues, termed ``scaffolds", around one or more subsections of a (``motif") protein structure that have known biologically-important functions through its interaction with a target. This enables the design of proteins with \emph{a priori} functional sites using generative models \citep{wang2021deep,watson_novo_2023}. The motifs can be small and have non-specific shapes (e.g. a helix), and hence it is important for the generative model to understand the chemical information it carries on top of its geometry. We thus consider the task of motif scaffolding as an example of how our model can be fine-tuned for conditional generation tasks. 
We consider two datasets for evaluating motif scaffolding performance: the benchmark proposed in~\citet{watson_novo_2023} consisting of $24$ single-chain motifs, and a new benchmark based on scaffolding the Complimentary Determining Regions (CDRs) of VHH nanobodies, as found in the Structural Antibody Database~\citep{schneider2022sabdab}.

\looseness=-1
\xhdr{Motif Scaffolding Benchmark}
We use the scaffolding benchmark from~\citet{watson_novo_2023} and follow the pseudo-label fine-tuning procedure described in~\citet{yim2023se} by randomly generating motifs from proteins by training on both the motif structure \emph{and} sequence. For inference, we sample the scaffold lengths for each motif and provide both the both partially masked structure and sequence to the model. We follow the same evaluation procedure used in RFDiffusion (c.f.~\S\ref{app:motif} for details). Our results in \cref{tab:combined_motif_scaffolding} show that both \foldflowpp and RFDiffusion solve all $24/24$ motifs.

\looseness=-1
\xhdr{CDR Scaffolding}
VHH antibodies, also known as nanobodies, have shown significant promise in protein design and therapeutics due to their unique properties~\citep{muyldermans2021applications}. They are composed of a single variable domain derived from camelid heavy-chain antibodies, featuring three complementarity-determining regions (CDRs) that confer specificity and variability in antigen binding. As a result, creating effective scaffolds for nanobodies is challenging due to the need to maintain the designability of the CDRs and especially because any scaffolding effort must avoid altering these characteristics to preserve binding functionality.
We treat this as a conditional generative modeling problem and fix the motif atoms, and mask the scaffold sequence information. Exact training and experimental details along with additional metrics are provided in~\S\ref{app:motif}. Our results are found in ~\cref{tab:combined_motif_scaffolding}, where the average motif $\rm scRMSD$ is much higher than the average scaffold $\rm scRMSD$. The result is a much lower number of solved motif scaffolding.

\begin{table}[H]
    \label{tab:motif}
    \centering
    \footnotesize
    \caption{\small Motif-scaffolding benchmarks. FrameFlow does not have public code for motif-scaffolding and thus cannot be evaluated on the VHH benchmark. ``+FT'' indicates ``with fine-tuning''. *Using reported numbers with AlphaFold2 instead of ESMFold used in our evaluation procedure; c.f.\ \S\ref{app:motif} for further discussion.}
    \begin{tabular}{lcc|ccc}
        \toprule
        Benchmark & \multicolumn{1}{C{2.5cm}}{RFDiffusion} & \multicolumn{3}{c}{VHH} \\
        \midrule
        Model & Solved /24 $\uparrow$ & Diversity $\uparrow$ & Motif $\downarrow$ & Scaffold $\downarrow$ & Solved /25 $\uparrow$ \\
        \midrule
        RFDiffusion  & 24 & 0.345 &  3.94 $\pm$ 1.54 & 2.40 $\pm$ 0.93 & 5 \\
        FrameFlow (+FT)* & 21 & -- & -- & -- & -- \\
        \foldflowpp (+FT) & 24 & 0.445 & 2.78 $\pm$ 1.01 & 1.67 $\pm$ 0.24 & 9 \\
        \bottomrule
    \end{tabular}
    \label{tab:combined_motif_scaffolding}
\end{table}
\vspace{-5mm}
\subsection{Zero-shot Equilibrium Conformation Sampling}
\label{sec:conformation_sampling}
\begin{wrapfigure}{r}{0.5\textwidth}
    \label{fig:md_samples} 
        \centering
        \includegraphics[width=0.4\textwidth]{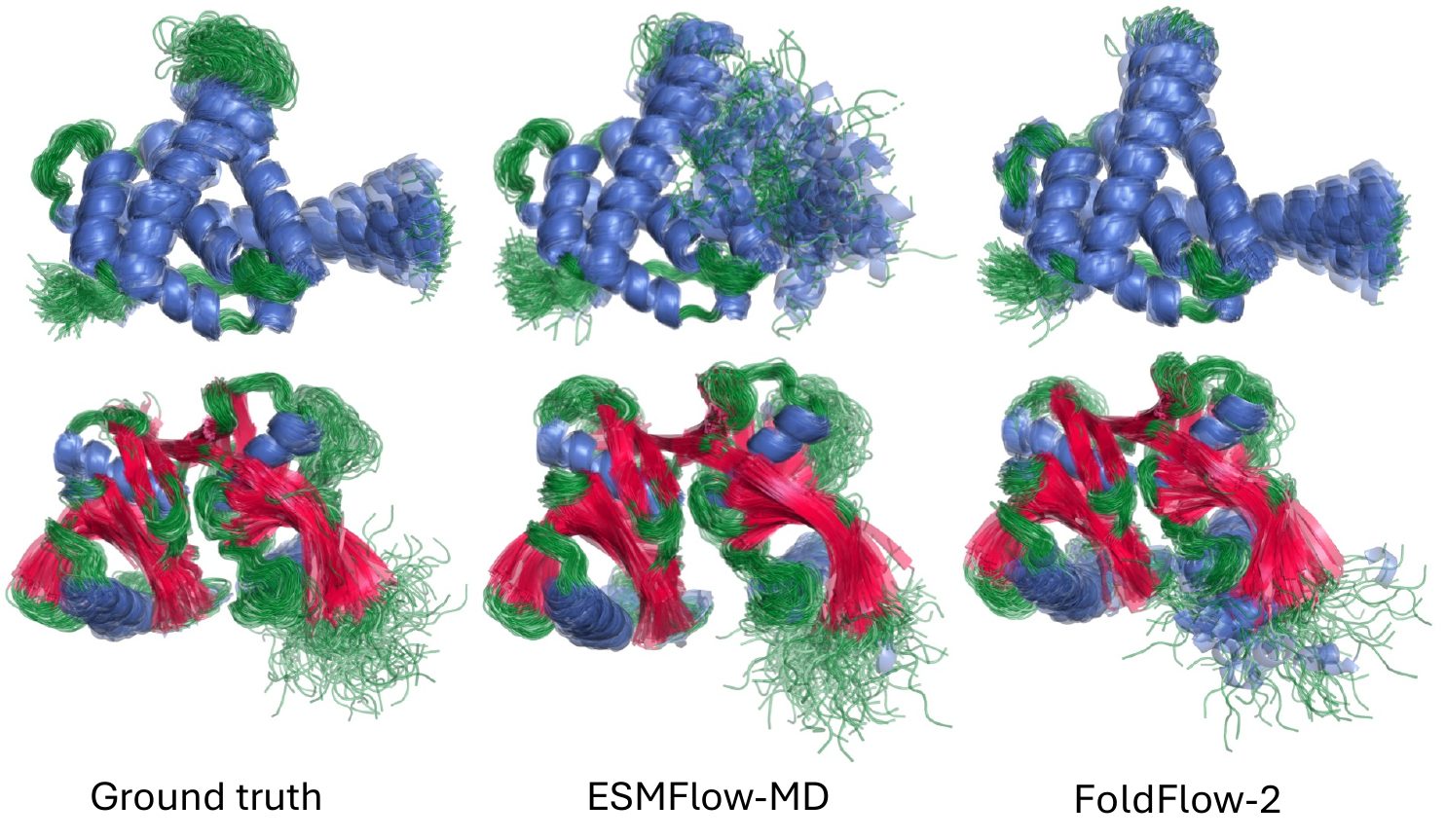}
        \caption{\small \looseness=-1 Protein conformation ensembles from the ATLAS dataset, ESMFlow-MD and \foldflowpp. Proteins are colored by their secondary structure with $\alpha$-helices in blue, $\beta$-sheets in red, and coils in green.}
\end{wrapfigure}
\looseness=-1
We now test \foldflowpp on zero-shot equilibrium conformation sampling task. Starting from a sequence, we generate multiple conformations of the same proteins and compare the distribution of conformations with the ones from molecular dynamic simulations. We compare \foldflowpp with AlphaFlow-MD and ESMFlow-MD; two folding models fine-tuned on a molecular dynamic dataset, and non finetuned models Eigenfold~\citep{jing2023eigenfold} and STR2STR~\citep{lustr2str}. In \cref{tab: md-task}, we report the pairwise and global RMSD, the root mean square fluctuation (RMSF), and the 2-Wasserstein on the top two principal components. For both RMSD and the RMSF metrics, we report the Pearson correlation between the values from the generated ensemble and those of the ground truth ensemble (the procedure is detailed in full in~\S\ref{app:conformation_sampling}).

We use the same test set as in~\citet{jing2024alphafold} restricted to proteins of length at most $400$ amino acids. Notably \foldflowpp performs similarly or better than the comparable model ESMFlow-MD across all metrics \emph{without any fine-tuning} and \emph{with significantly fewer parameters} on molecular dynamics data, indicating that the base model trained only on PDB already captures similar information about protein dynamics as models given explicit access to this data. Moreover, we observed that \foldflowpp requires 4.5$\times$ less GPU hours for training and 33$\times$ less trainable parameters while allowing for 6$\times$ faster inference steps than ESMFlow-MD as reported in \cref{tab:md-speed}, improving \foldflowpp's prospects as a practical base model for future work on capturing protein dynamics.

\begin{table}[ht!]
    \caption{\small Zero-shot performance of the base \foldflowpp model on the ATLAS dataset of MD trajectories compared to ESMFlow and AlphaFlow models fine-tuned on ATLAS. \foldflowpp is competitive to the comparable model ESMFlow across all metrics. $r$ denotes Pearson's correlation coefficient. Time is per sample (Time / sample (s)) on length 300 protein 7c45\_A. Eigenfold only models C$\alpha$ atom so we compare on PCA $\mathcal{W}_2$-dist on the C$\alpha$ atom only.}\label{tab: md-task} 
    \centering
    \resizebox{\textwidth}{!}{
    \begin{tabular}{lrrrrrrr}
        \toprule
        Model & Pairwise RMSD $r$ ($\uparrow$) & Global RMSF $r$ ($\uparrow$) & Per-target RMSF $r$ ($\uparrow$) & PCA $\mathcal{W}_2$-dist ($\downarrow$) &  PCA $\mathcal{W}_2$-dist (C$\alpha$ only) ($\downarrow$) &   Time / sample (s) \\ %
        \midrule
        AlphaFlow-MD  & 0.468 $\pm$ 0.005 & 0.415 $\pm$ 0.006 & 0.824 $\pm$ 0.000 & 10.67 $\pm$ 0.29  & --- &  32.6 $\pm$ 0.1\\ %
        ESMFlow-MD  & 0.293 $\pm$ 0.005 & 0.161 $\pm$ 0.005 & 0.737 $\pm$ 0.001 & 11.51 $\pm$ 0.13 & --- &  11.2 $\pm$ 0.1 \\ %
        \midrule
        \foldflowpp  & 0.297 $\pm$ 0.004 & 0.236 $\pm$ 0.004 & 0.658 $\pm$ 0.001 & 10.85 $\pm$ 0.15 &  5.273 $\pm$ 0.075 &  9.0 $\pm$ 0.0 \\ %
         STR2STR &  0.279 $\pm$ 0.003	 &  0.244 $\pm$ 0.005 & 	 0.586 $\pm$ 0.001	&  5.37 $\pm$ 0.05 &  2.678 $\pm$ 0.023  &  10.6 $\pm$ --- \\
         Eigenfold &  0.194 $\pm$ 0.004 & 	 0.156	$\pm$ 0.004 & 0.668 $\pm$ 0.001	& --- & 5.296 $\pm$ 0.080  & 38.6 $\pm$ 0.5\\ %
        \bottomrule
    \end{tabular}
    }
\end{table}

\section{Related work}
\label{sec:related_work}
\looseness=-1
\xhdr{Protein design} Physics-based protein structure design yielded the first de novo proteins~\citep{huang2016coming}. For example, structure-based biophysics approaches have previously resulted in several drug candidates~\citep{rothlisberger2008kemp,fleishman2011computational,cao2020novo}.This was followed by language models~\citep{hie2022high, ferruz2022protgpt2} and geometric deep learning~\citep{gainza2020deciphering} for protein structure design. Recently, diffusion~\citep{wu2024protein,yim2023se,watson_novo_2023,ingraham2023illuminating,wang2024proteus,frey2023protein} and flow-based models~\citep{bose2023se3stochastic,yim2023fast, jing2024alphafold} have risen to prominence.
These methods employ a backbone-first approach with the notable exception of MultiFlow~\citep{campbell2024generative} which uses sequence to perform co-generation.

\looseness=-1
\xhdr{RLHF and Supervised Fine-Tuning (SFT)} Aligning the outputs of language models with RLHF has recently gained interest \citep{Ouyang0JAWMZASR22, StiennonO0ZLVRA20, BaiRLHF}. These methods learn a reward model for post-training alignment to desired behavior \citep{MishraKBH22}, which can prove challenging for protein design~\citep{EnergyDPO}. SFT on hand-crafted data has proven to be effective in enhancing performance but requires high-quality data \citep{SFTCodeGen, SFTYuan}. Filtering real data using auxiliary rewards serve as a substitute for steering the desired properties of the generated samples.

\section{Conclusion}\label{sec:conclusion}
\looseness=-1
In this paper, we introduce a new sequence-conditioned protein structure generative model called \foldflowpp. \foldflowpp leverages a protein language model to condition Flow Matching-based protein generative models with sequences. Our model achieves state-of-the-art results on unconditional generation and generates diverse and novel proteins, especially when trained on our new dataset. Conditioning over sequences allows our model to perform novel tasks such as folding sequences and motif-scaffolding tasks and we show its competitiveness on those tasks. Regarding the limitations of our model, we note that it requires a competitive pre-trained language model to be sequence conditioned, which can be hard to acquire. We also note that ProteinMPNN, used in our evaluation pipeline, has been trained only on PDBs. Therefore, it is possible that our models trained on our new dataset generates designable proteins which are not correctly processed by ProteinMPNN.

\section*{Acknowledgements}
We thank Alexandre Stein, Maksym Korablyov, and the entire DreamFold team for providing a vibrant workspace that enabled this research. The authors would like to acknowledge Anyscale, Amazon AWS, and Google GCP for providing computational resources for the protein experiments.

\section*{Contribution statement}
Architecture design was led by G.H. Infrastructure development was led by J.V. The experiments were divided as follows: Unconditional (K.F., A.T.), diversity (E.T.L., R.I., C.L.), folding (G.H.), motif-scaffolding (G.H., J.V., E.T.L, C.L.), and equilibrium conformation (J.R.B, G.H., T.A.S.). Dataset preparation including synthetic structure filtering (J.V., P.L., K.F.). A.J.B. drove the writing of the paper with contributions from all other authors. A.J.B. and A.T. cosupervised the project with guidance from M.B.

\bibliographystyle{abbrvnat}
\bibliography{tidy_bibliography}

\clearpage
\appendix

\section*{Broader impact}

The development of generative AI for protein backbone design holds significant promise for the fields of biotechnology and medicine. By enabling the precise engineering of protein structures, these models can accelerate the discovery of novel therapeutics and vaccines, potentially leading to more effective treatments for a wide range of diseases. Multi-modal and conditional generative models in the space of biotechnology are seeing rapid improvements, from which we have yet to see the full impact on modern and future medicine. We also acknowledge the potential for dual use of our results but, as discussed in \citet{IPDresponsibleAI}, the benefits of public research into computational drug discovery outweigh the potential drawbacks.

\section{A short review of Riemannian geometry, Lie groups, and optimal transport}
\label{app:riemannian_geo_primer}

\subsection{Riemannian manifolds} 
Informally, a {\em topological manifold}, $\gM$ is a topological space that is locally Euclidean (i.e. homeomorphic to a Euclidean space). The manifold is said to be {\em smooth} or {\em differentiable} if it additionally is $C^p$ differential for all $p$. An important notion is the {\em tangent space}, $\gT_x \gM$, which is attached to every point on the manifold, $x\in \gM$. The disjoint union of all the tangent spaces is called a {\em tangent bundle}, $\mathfrak{X}(\gM)$. If in addition, the manifold is equipped with a 
{\em Riemannian metric}, $g_x$, it is said to be a {\em Riemannian manifold}. The notion of a Riemannian metric is used to define inner products on the tangent space at each point of the manifold. This means that for $\mathfrak{x_1}, \mathfrak{x_2} \in \gT_x \gM$, $g_x(\mathfrak{x_1}, \mathfrak{x_2}) := \langle \mathfrak{x_1}, \mathfrak{x_2} \rangle$. Similar to how inner products can be used to define key geometric properties, such as length and distance, the Riemannian metric allows us to define such notions on an arbitrary Riemannian manifold: the length of $\mathfrak{x}$, a vector in the tangent space of the manifold is defined by $|\mathfrak{x}| = \langle \mathfrak{x}, \mathfrak{x} \rangle ^{1/2}$. Finally, an important property on Riemannian manifolds is {\em geodesics}, which generalizes the notion of shortest paths in Euclidean spaces. While in Euclidean space the shortest path between two points is the length of a straight line between them, on a manifold, the idea is to find the shortest smooth curve between two points, which is called a geodesic.

\cut{
an $n$-dimensional {\em manifold} $\gM$ is a topological space locally equivalent (homeomorphic) to $\mathbb{R}^n$. 
This implies that one has the notion of `neighbourhood' but not of `distance' or `angle' on $\gM$. The manifold is said to be {\em smooth} if it additionally has a $C^\infty$ differential structure. 
At every point $x\in \gM$, one can attach a {\em tangent space} $\gT_x$. The disjoint union of tangent spaces forms the {\em tangent bundle}. 
A {\em Riemannian manifold}\footnote{We tacitly assume $\gM$ to be orientable, connected, and complete and admit a volume form denoted as $dx$. } $(\gM, g)$ is additionally equipped with an inner product ({\em Riemannian metric}) $g_x: \gT_x\gM \times \gT_x\gM\rightarrow\R$ on the
tangent space $\gT_x\gM$ at each $x\in\gM$. The Riemannian metric $g$ allows to define key geometric quantities on $\gM$ such as distances, volumes, angles, and length minimizing curves ({\em geodesics}). 
We consider functions defined on $\gM$ and the tangent bundle, referred to as {\em scalar-} and {\em vector fields}, respectively. The {\em Riemannian gradient} is an operator $\nabla_g: C^\infty(\gM) \rightarrow \mathfrak{X}(\gM)$ between the respective functional spaces. Given a smooth scalar field $f\in C^\infty(\gM)$, its gradient $\nabla_g f \in \mathfrak{X}(\gM)$ is the local direction of its steepest change. 
}

\subsection{Lie groups}
\looseness=-1
Symmetries refer to transformations of an object that preserve a certain structure. A set of continuous symmetries, paired with a composition operation satisfying group axioms is a {\em Lie group} $(G, \circ)$. More precisely, a group is a set paired with a group operation, $\circ: G \times G \to G$, which is associative, has an identity element and there is an inverse element for every element of the set. In addition to this group structure, a Lie group is also a smooth manifold, where the group operations of multiplication, $(x, y) \rightarrow xy$ for $x, y \in G$, and inversion, $x \rightarrow x^{-1}$, are smooth maps.  

Given $y \in G$, we can define a diffeomorphism, $L_y: G \to G$ defined by $x \mapsto yx$, known as left multiplication. Given a vector field $X$ on the group, we say that it is {\em left invariant}, if, under this left multiplication, it is left-invariant, meaning that $L_y^* X = X$, $\forall y \in G$. Note that $L_y^*$ is the differential of left action, naturally identifying the tangent spaces, $\gT_y \rightarrow \gT_{yx}$. Therefore, given the group multiplication, we can {\em uniquely} define a left-invariant vector field with its values on the tangent space at the identity element of the group, $\gT_e$. We can additionally equip this vector space, $V$, with a bilinear operation known as the {\em Lie bracket}, $[ \cdot, \cdot ]: V \times V \rightarrow V$, that satisfies the Jacobi identity and is anticommutative. Such a vector space is called a {\em Lie algebra}. The tangent space of Lie groups form Lie algebras and are denoted with $\mathfrak{G}$. The elements of the Lie algebra can be mapped into group elements using an invertible map called the {\em exponential map}, $\exp: \mathfrak{G} \rightarrow G$. The inverse is called the {\em logarithmic map} allowing us to go from the group elements to their corresponding elements in the Lie algebra, $\log: G \to \mathfrak{G}$.

\looseness=-1
Finally, we note that the set of $n \times n$ non-singular matrices forms a Lie group. The group operation is matrix multiplication and it can be seen as a smooth manifold that is an open subset of $\R^{2n}$. This group is known as the {\em General Linear Group}, $GL(n)$. Any closed subgroup of $GL(n)$ is known as a {\em matrix Lie group}, which are perhaps some of the most important Lie groups that are studied. In the case of these matrix Lie groups, the exponential and logarithmic maps also coincide with the matrix exponential and logarithm. For a more detailed overview of this subject, we refer the reader to \cite{hall2013lie}.

\subsection{The Special Euclidean Group in 3 Dimensions}
\looseness=-1
One of the closed subgroups of $GL(n)$ that has been studied extensively in various fields is the $3D$ Special Orthogonal group, $\sothree$. The elements of this group are $3 \times 3$ rotation matrices, namely $SO(n) = \{ r \in GL(n) : r^Tr=I, det(r)=1 \}$. Additionally, the translations of an object in $3D$ space by a translation vector $s$ can also be seen as a matrix Lie group by considering the translation matrix, 
$\begin{pmatrix}
I & s \\
0 & 1
\end{pmatrix} $, where $I$ is an $3 \times 3$ identity matrix. With the group operation being translations, this group is also a matrix group, denoted as $(\R^3, +)$.

Combining these, we can represent the {\em rigid transformations} of objects in $3D$ space with a group that encompasses both rotations and translations. This group is known as the {\em Special Euclidean group in $3D$} and is the semidirect product of the rotation and translation groups, $\sethree \cong \sothree \ltimes (\R^3, +)$. This group is also a matrix group and its elements can be written as $\sethree = \big \{ (r, s) = \begin{pmatrix}r & s \\ 0 & 1  \end{pmatrix}: r \in \sothree, s \in (\R^3, +) \big \}$. 
Finally, we note that given a suitable choice of metric~\citep{park1994kinematic} the inner product on $\sethree$ decomposes naturally into inner products on $\sothree$ and $(\R^3, +)$, i.e. $\langle \mathfrak{x}_1, \mathfrak{x}_2 \rangle_{\sethree}  = \langle \mathfrak{r}_1, \mathfrak{r}_2 \rangle_{\sothree} + \langle s_1, s_2 \rangle_{(\R^3, +)}$, where we have denoted the elements of tangent spaces of $\sothree$ and $\R^3$ with $\mathfrak{x}$ and $\mathfrak{r}$ respectively and have omitted to do so for the translation group as the tangent space coincides with the space itself.

\subsection{Optimal Transport}
Optimal transport (OT) focuses on finding the most efficient way to move mass or resources from one distribution to another. It involves minimizing a cost function \( c(x, y) \) that quantifies the expense of transporting mass from point \( x \) in one space to point \( y \) in another space. This problem is framed as an optimization problem, often using the Kantorovich formulation:

\[
\min_{\gamma \in \Pi(\mu, \nu)} \int_{X \times Y} c(x, y) \, d\gamma(x, y)
\]

where \( \Pi(\mu, \nu) \) is the set of all joint distributions \(\gamma\) with marginals \(\mu\) and \(\nu\). The goal is to find a transportation plan \(\gamma\) that minimizes the total cost. Optimal transport has significant applications in probability theory, where it is used to compare probability distributions. One common measure is the Wasserstein distance, defined as:

\[
W_p(\mu, \nu) = \left( \inf_{\gamma \in \Pi(\mu, \nu)} \int_{X \times Y} d(x, y)^p \, d\gamma(x, y) \right)^{1/p}
\]

where \( d(x, y) \) represents the distance between points \( x \) and \( y \) and \( p \geq 1 \) is a parameter that defines the type of Wasserstein distance.

In machine learning, optimal transport enhances algorithms in tasks such as domain adaptation, where it helps align different data distributions, and in generative modeling, where it aids in generating data samples that match a given distribution. In the case of \foldflowpp and as used in \citep{bose2023se3stochastic}, OT is used to sample tuples of source noisy sample and target protein $(x_0, x_1)$ to perform Flow Matching. It takes the form of using the OT plan for the joint distribution $q(x_0, x_1)$ between minibatches $(X_0, X_1)$. The OT variant performs here corresponds to what is called minibatch OT as studied in \citep{fatras2021minibatch, fatras20a}.

\section{Experimental Details}
\label{app:experimental_details}

\subsection{Dataset filtering}
\label{app:dataset_filtering}

\subsubsection{PDB Structures}
We use a subset of PDB with resolution $< 5 \angstrom$ downloaded from the PDB~\citep{berman_protein_2000} on July 20, 2023. We performed standard filtering to remove any proteins with > 50\% loops. During preprocessing, we also removed any non-organic residues at either end of the structure. In previous works, these residues are typically kept but masked during training, however they contributed to the total forward pass FLOPs and therefore decrease training efficiency. By removing these residues during preprocessing, we are able to increase the number of training examples per batch. Finally, we re-clustered the PDB dataset using \texttt{mmseqs2} at the 50\% sequence identity threshold to obtain 6,593 clusters during training. The PDB is made available under a CC0 1.0 Universal (CC0 1.0) Public Domain Dedication.

\subsubsection{Synthetic Structures}
In this section, we provide more detail on the filtering procedure used in our curation of synthetic data for training. We began with a SwissProt data dump consistent of 532,003 structures predicted by AlphaFold2~\citep{jumper2021highly} accessed in February 2024. The AlphaFold2 predicted structure database is made available under a CC-BY-4.0 license for academic and commercial uses. 

\begin{figure}[ht!]
    \begin{tabular}{c c}
        \includegraphics[scale=0.3]{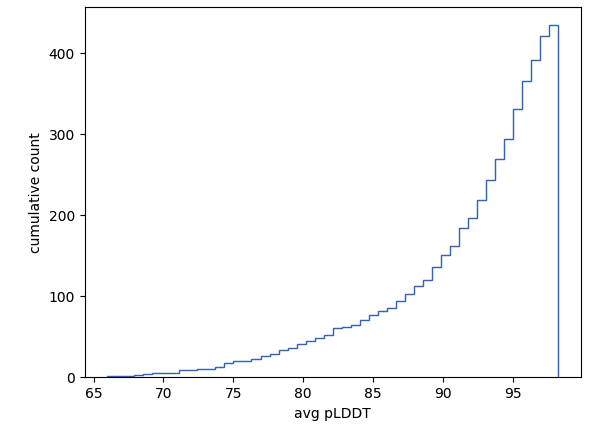} & 
        \includegraphics[scale=0.3]{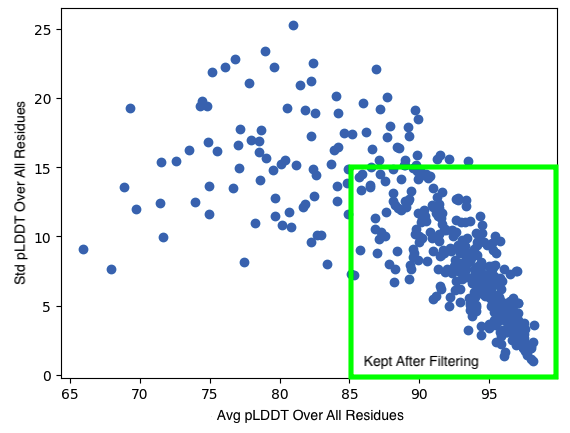} \\
        \textbf{(a)} & \textbf{(b)}
    \end{tabular}
    \caption{Analysis of global pLDDT distribution on a sample of 500 proteins from SwissProt.}
    \label{fig:swissprot_analysis}
\end{figure}

\xhdr{Global pLDDT Filtering}
We first filtered out globally low-confident structures, as measured by average and standard deviation of pLDDT taken across all residues in a predicted structure. Despite SwissProt already being a curated set of high-confidence structures, we still found considerable variation in quality. See \cref{fig:swissprot_analysis}a for an empirical analysis average pLDDT for a random sample of 500 proteins from SwissProt. Our final filtering criteria were \texttt{(avg pLDDT > 85) \&\& (std pLDDT < 15)} to keep only ``consistently good'' structures, see \cref{fig:swissprot_analysis}b for a graphical representation.

\begin{figure}[ht!]
    \centering
    \begin{tabular}{c c c}
        \includegraphics[scale=0.35]{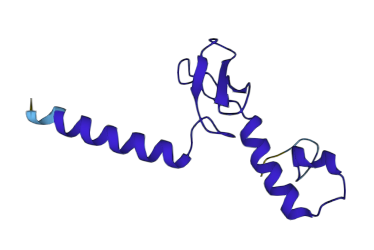} & \includegraphics[scale=0.35]{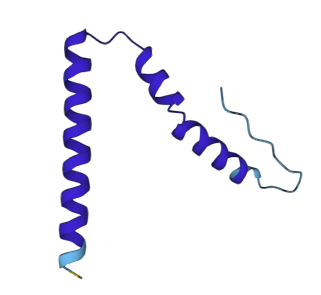} & 
        \includegraphics[scale=0.35]{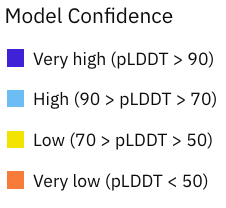} \\
        \texttt{A4FZT8} & \texttt{Q0SNQ5}
    \end{tabular}
    \caption{Examples of high-confidence, low-quality structures that were filtered out of the training set. Images are from the AlphaFold Protein Structure Database \url{https://alphafold.ebi.ac.uk/} accessed in May 2024; identifiers are the UniProt IDs of each example.}
    \label{fig:high_conf_low_qual}
\end{figure}

\xhdr{High-Confidence Low-Quality Filtering}
Despite the global pLDDT filtering, there were still low-quality structures in the training after global pLDDT filtering. Some examples of these structures can be seen in \cref{fig:high_conf_low_qual}. They are characterized by having high overall confidence and good local qualities but unrealistic global structures or sub-chain interactions. Our finding is that these structures easily corrupt the training data and cause a model to produce similarly ``unfolded'' generations. 

\subsection{Model architecture details}
We provide details for the IPA blocks and Folding blocks. For the IPA blocks, we follow the setting developed in~\citep{yim2023se} by adding to the original IPA~\citep{jumper2021highly} a skip connection and transformer layer on the node representation. 
The IPA modules takes as inputs the single and pair representations and a structure. For the structure encoder, we initialize the single and pair representations with positional and time embedding passed to MLPs. In our experiment, we used a node embedding dimension of 256, an edge dimension of 128, and a hidden dimension of 256. These settings are the same for both the encoder and decoder. We have use a skip connection between the representations of the structure encoder and decoder. 

We combine the single and pair representations of different modalities by projecting each with a linear layer of output dimensions 128 for the single representations and 64 for the pair representation. We then concatenate all modalities' representation to obtain a single representation of dimension 128 and pair representation of dimension 256.

The Folding blocks are taken from \cite{lin2022language}. They are composed of 2 Triangular Self-Attention Blocks with single and pair head width of size 32. Finally, the pair and single representation dimensions are of 128. 

The structure decoder's single and pair representations inputs are an average between the refined ones from the Folding blocks and the initial ones. All other architectural details not specified here are set to the defaults of IPA or ESM, respectively.

\subsection{Training Details}\label{app:training details}

\xhdr{Hyperparameters} See \cref{tab:training} for an overview of the experimental setup. We train with the ``length batching'' scheme described in \cite{yim2023fast} in which each batch consists of the same protein sampled at different times. The number of samples in a batch is variable and is approximately $\lceil \mathrm{num\_residues}^2 / M\rceil$ where $M$ is a hyperparameter in \cref{tab:training}. Other training details such as the loss computation are the same as \citet{bose2023se3stochastic}. 

\xhdr{Training hardware setup} \foldflowpp is coded in PyTorch and was trained on 2 A100 40GB NVIDIA GPUs for 4 days. Initial tests runs were trained in a similar setting.

\begin{table}
    \centering
    \caption{Overview of Training Setup}
    \label{tab:training}
    \begin{tabular}{l c}
        \toprule
        Training Parameter & Value \\
        \midrule
        Optimizer & ADAM \cite{kingma2014adam} \\
        Learning Rate & 0.0001 \\
        $\beta_1,~\beta_2,\varepsilon$ & 0.9, 0.999, 1e-8 \\
        Effective $M$ (max squared residues per batch) & 500k \\
        \% of experimental structures per epoch & 33\%\\
        Minimum number of residues & 60 \\
        Maximum number of residues & 384 \\
        Sequence masking probability & 50\% \\
        \bottomrule
    \end{tabular}
\end{table}

\begin{table}
    \centering
    \label{tab:training_time}
    \caption{An overview of training time. *RFDiffusion initializes from RoseTTAFold, and we include that training time in the estimates. **We recall that \foldflowpp uses frozen ESM2-650M which was trained on 512 GPUs for 8 days.}
    \begin{tabular}{l r r r r}
        \toprule
        Model & \# Steps & \# GPUs & Total Time (days) \\
        \midrule
        RFDiffusion* \citep{watson_novo_2023} & 25k &  64 + 8 & 28 + 3\\
        \foldflow \citep{bose2023se3stochastic} & 600k & 4 & 2.5 \\
        \foldflowpp** & 500k & 2 & 4 \\
        \foldflowpp** w/o folding block & 500k & 2 & 2.5\\        
        \bottomrule
    \end{tabular}
\end{table}

\begin{figure}
    \centering
    \includegraphics[width=1.0\textwidth]{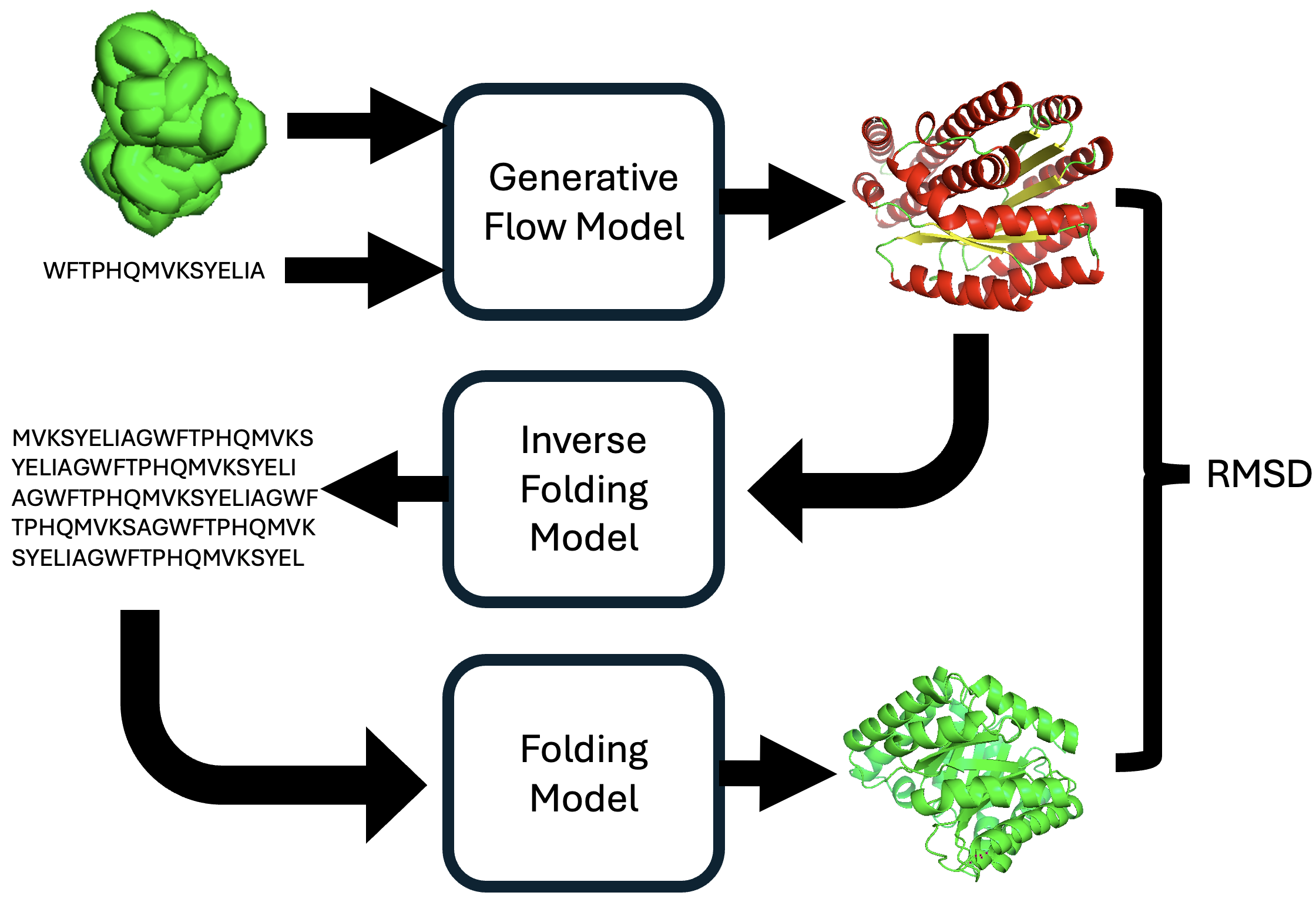}
    \caption{Schematic of designability calculation. First a generative flow model is used to generate a protein backbone from an initial structure (possibly noise) and (optionally) a protein sequence. This is then fed to an inverse folding model (ProteinMPNN~\cite{dauparas_robust_2022}) eight times to generate eight sequences for the structure. Then all eight sequences are fed back to ESMFold to produce a structure for each sequence. All eight structures are compared using scRMSD from the ESMFold ``refolded'' structure to the generated structure, and the minimum is taken as the ``designability'' of the generated structure with at least one structure with error $< 2.0 \angstrom$ being classified as designable following~\cite{watson_novo_2023}.}
    \label{fig:refold-schematic}
\end{figure}

\subsection{Inference details}
\label{app:inference_details}
The inference is performed with Euler integration steps. In our experiments, we found that 50 steps gave state-of-the-art results. We use the \textit{Inference Annealing} trick from \citet{bose2023se3stochastic} multiplying the rotation vector by some time dependent scaling function $i(t)$ where $t \in [0, 1]$. We tried different scaling parameters and as found in \citep{bose2023se3stochastic}, $10t$ provided the best performance. In practice this means greatly speeding up the rotation components at the beginning of inference ($t\approx1$) and slowing it down at the end of inference ($t\approx 0$).

\subsection{Unconditional protein backbone generation}

\xhdr{Metrics}
We compute several quantity of interest to measure the performance of \foldflowpp. i) We start with designability. We measure designability using the \emph{self-consistency} metric with ProteinMPNN~\citep{dauparas_robust_2022} and ESMFold~\citep{lin2022language}, counting the fraction of proteins that refold ($C_\alpha$-RMSD (scRMSD) < $2.0\angstrom$) over $50$ proteins at lengths $\{100, 150, 200, 250, 300\}$. For our ablation study and sensitivity analysis in \S~\ref{sec:ablation}, we also provide the average self-consistency rmsd over all generated proteins. 

ii) In order to use generative models for drug discovery applications, we want to measure how different and novel are the generated data compared to training data. We measure novelty using two metrics: 1.) the fraction of generated proteins that are both designable and are dissimilar to PDB structures (quantified by template-match score to PDB, \emph{i.e.,} PDB-TM score) $<0.3$, as used in~\citet{lin2023generating}, higher is better) and 2.) for designable proteins, the average closest similarity to training data (quantified by the maximum TM score, lower is better). We note that the threshold for similarity has been studied previously, where the average TM-score on random structure pairs is $\sim 0.3$~\citep{zhang2006origin}.

iii) Finally we want a model that generates diverse proteins and not just of the same type. Proteins are usually gathered into different clusters during training. So for diversity, we use the number of generated clusters with a TM-score threshold of 0.5~\citep{herbert2008maxcluster}(higher is better) as well as the average pairwise TM-score of the \textit{designable} generated samples averaged across lengths as our diversity metric (lower is better). Note that in certain model, designability is inversely correlated with diversity as these models can produce unrealistic (e.g.\ unfolded) proteins that are ``diverse'' because they do not align well with each other. 

\xhdr{Baselines}
On the unconditional backbone generation task we compare to pre-trained versions of FrameDiff~\citep{yim2023se}, Chroma~\citep{ingraham2023illuminating}, Genie~\citep{lin2023generating}, FoldFlow~\citep{bose2023se3stochastic}, and RFDiffusion~\citep{watson_novo_2023}. We use the default parameters for each model including \# of Euler steps for inference and default noise levels (0.1 for RFDiffusion and FrameDiff). We use the OT version of FoldFlow as it is the most similar to our setup and achieved the highest designability.

In \cref{tab:main_table} we use 30 Euler steps for inference to better match the diversity levels of the baseline models for more accurate comparison on these metrics. Results for the 50 step model can be seen in \cref{tab:euler_steps_unconditional_gen} with slightly worse designability but improved novelty and diversity.

\subsection{Motif scaffolding}\label{app:motif}
\begin{figure}
    \centering
    \caption{Samples of solved motif scaffolding problems from the benchmark of \citet{watson_novo_2023}. The motif is in red, the designed scaffold is in blue, and the refolded structure from ESMFold is in gray.}
    \label{fig:motif_samples}
    \begin{tabular}{c c c}
        \includegraphics[width=4cm]{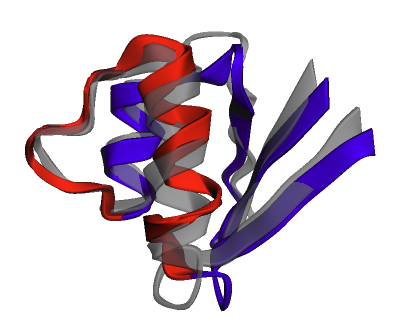} & 
        \includegraphics[width=3.5cm]{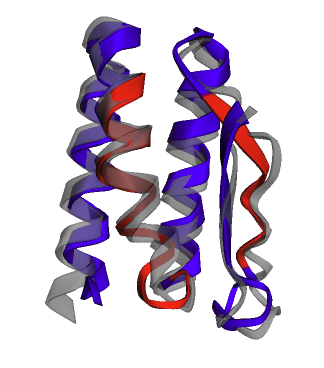} & 
        \includegraphics[width=3.5cm]{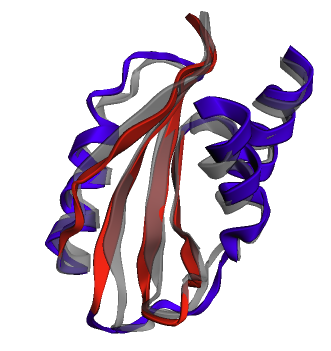} \\
        \texttt{7MRX\_med} & \texttt{4JHW} & \texttt{5IUS} \\
        \includegraphics[width=3.5cm]{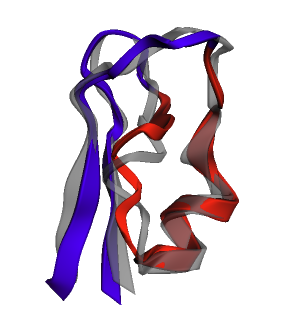} & 
        \includegraphics[width=3.5cm]{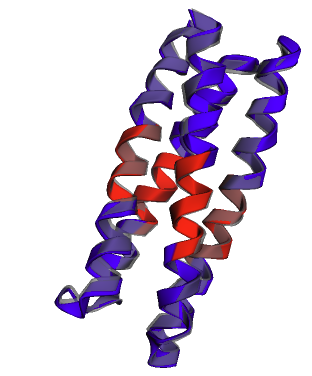} & 
        \includegraphics[width=3.5cm]{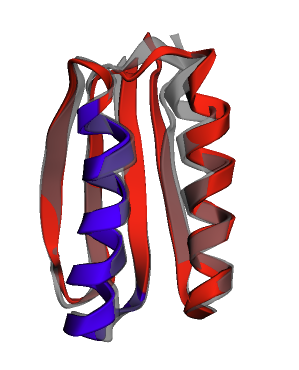} \\
        \texttt{5WN9} & \texttt{1BCF} & \texttt{2KL8} \\
    \end{tabular}
\end{figure}

\xhdr{Evaluation procedure} We follow the same evaluation procedure as in \cite{watson_novo_2023, yim2023se}. In particular, to evaluate the designability of a scaffold, we use ProteinMPNN \citep{dauparas_robust_2022} to decode 8 sequences and then re-fold those sequences, \emph{fixing the motif sequence} which is known a priori. Given these re-folded structures, we compare three numbers:
\begin{enumerate}
    \item \textbf{Global RMSD}: the overall aligned RMSD between the entire generated structure and the refolded structure.
    \item \textbf{Motif RSMD}: the RMSD of the refolded motif residues aligned to the original motif residues.
    \item \textbf{Scaffold RMSD}: the RMSD of the refolded scaffold residues aligned to the generated scaffold residues.
\end{enumerate}
Following \cite{watson_novo_2023}, a scaffold is considered ``designable'' if the Global RMSD is $<2$ AND the motif RMSD $<1$ AND the scaffold RMSD $<2$. A detailed breakdown of our results can be found in \cref{tab:motif_detailed}, with some samples of designable motifs in \cref{fig:motif_samples}. 

We note that the choice of folding model appears to have a nontrivial impact on this metric. In their original papers, \citet{watson_novo_2023,yim2023se} used AlphaFold2 with no MSA and 0 recycles to refold their structures; however ESMFold is known to be significantly more accurate when no MSAs are provided \citep{lin2023evolutionary}. Given this, we generated new samples from RFDiffusion (FrameFlow doesn't have public code for generating scaffolds) and re-folded them with ESMFold. The result is that RFDiffusion is able to solve all 24 examples; an increase of 4 vs. their reported numbers. Moreover, the proportion of solved increases relative to their reported results, suggesting that the accuracy of the folding model significantly impacts the ability to measure scaffold quality \emph{in silico}. 

\begin{table}
    \caption{A detailed breakdown of \foldflowpp motif scaffolding performance using ESMFold to refold all structures. All numbers are out of 100 samples.}
    \label{tab:motif_detailed}
    \resizebox{1.0\textwidth}{!}{
    \begin{tabular}{l|rrr|r|r}
        \toprule
        & \multicolumn{4}{c|}{\foldflowpp} & RFDiffusion \\
        Example & \# Overall Valid & \# Motif Valid & \# Scaffold Valid & \# Designable & \# Designable\\
        \midrule
        1BCF & 100 & 100 & 100 & \textbf{100} & \textbf{100} \\
        2KL8 & 100 & 100 & 100 & \textbf{100} & \textbf{100} \\
        1PRW & 100 & 100 & 98 & \textbf{98} & 91 \\
        1YCR & 96 & 100 & 89 & 85 & \textbf{91} \\
        4ZYP & 89 & 97 & 87 & 80 & \textbf{85} \\
        3IXT & 97 & 101 & 75 & 73 & \textbf{85} \\
        7MRX\_small & 70 & 92 & 83 & \textbf{66} & 22 \\
        6EXZ\_long & 78 & 83 & 62 & 59 & \textbf{91} \\
        5TPN & 62 & 95 & 78 & 57 & \textbf{79} \\
        6EXZ\_med & 67 & 70 & 55 & 54 & \textbf{87} \\
        6EXZ\_small & 56 & 63 & 55 & \textbf{53} & 28 \\
        6E6R\_long & 84 & 98 & 57 & 51 & \textbf{82} \\
        1QJG & 54 & 82 & 76 & 49 & \textbf{80} \\
        7MRX\_med & 59 & 86 & 67 & \textbf{49} & 22 \\
        6E6R\_med & 78 & 96 & 54 & 43 & \textbf{87} \\
        5TRV\_med & 44 & 86 & 70 & 41 & \textbf{80} \\
        5TRV\_long & 40 & 85 & 69 & 38 & \textbf{77} \\
        6E6R\_small & 82 & 100 & 34 & 30 & \textbf{50} \\
        5TRV\_small & 28 & 78 & 47 & 24 & \textbf{53} \\
        7MRX\_long & 27 & 59 & 47 & 20 & \textbf{22} \\
        5YUI & 14 & 54 & 48 & \textbf{13} & 8 \\
        5IUS & 10 & 86 & 79 & 10 & \textbf{11} \\
        5WN9 & 6 & 12 & 10 & \textbf{4} & 1 \\
        4JHW & 3 & 74 & 39 & 3 & \textbf{8} \\
        \bottomrule
    \end{tabular}
    }
\end{table}

\begin{table}
    \caption{A detailed breakdown of \foldflowpp motif scaffolding performance applied to refoldable VHH structures. The same comments as \cref{tab:motif_detailed} apply to evaluation. All numbers are out of 25 samples.}
    \label{tab:motif_vhh_full}
    \begin{tabular}{l|rrr|r|r}
    \toprule
    & \multicolumn{4}{c|}{\foldflowpp} & RFDiffusion \\
    Example & \# Overall Valid & \# Motif Valid & \# Scaffold Valid & \# Designable & \# Designable\\
    \midrule
    6qtl-B & 4 & 21 & 16 & \textbf{4} & 0 \\
    6qtl-G & 4 & 19 & 18 & \textbf{4} & 0 \\
    6rpj-H & 4 & 21 & 21 & \textbf{4} & 0 \\
    6rpj-D & 3 & 23 & 19 & \textbf{3} & 0 \\
    6qtl-F & 2 & 22 & 18 & \textbf{2} & 0 \\
    7epb-C & 2 & 35 & 33 & 2 & \textbf{4} \\
    6rpj-F & 3 & 24 & 21 & \textbf{2} & 0 \\
    6qtl-C & 3 & 20 & 14 & \textbf{2} & 1 \\
    5l21-B & 1 & 16 & 7 & 1 & 1 \\
    6oz6-G & 0 & 16 & 2 & 0 & 0 \\
    6oz6-E & 0 & 17 & 1 & 0 & 0 \\
    6oyz-G & 0 & 12 & 0 & 0 & 0 \\
    6oyz-F & 0 & 15 & 0 & 0 & 0 \\
    6oyh-H & 0 & 17 & 0 & 0 & 0 \\
    6oyh-G & 0 & 19 & 1 & 0 & 0 \\
    6gs7-H & 1 & 19 & 9 & 0 & \textbf{2} \\
    1kxq-E & 0 & 23 & 13 & 0 & 0 \\
    6rpj-B & 0 & 21 & 19 & 0 & 0 \\
    7a50-C & 0 & 23 & 18 & 0 & 0 \\
    7epb-D & 0 & 22 & 19 & 0 & \textbf{2} \\
    7o31-X & 0 & 14 & 12 & 0 & 0 \\
    7q3q-B & 0 & 13 & 5 & 0 & 0 \\
    7tjc-B & 0 & 17 & 9 & 0 & 0 \\
    8cxr-E & 0 & 14 & 1 & 0 & 0 \\
    8cxr-F & 0 & 18 & 1 & 0 & 0 \\
    \bottomrule
    \end{tabular}
\end{table}

\subsubsection{Pseudo-label training} \label{app:pseudo-label}

We follow a similar data augmentation procedure as in \citep{yim2024improved,watson_novo_2023}, with the only modification of adding a minimum number of contiguous residues per motif. We use the same min length and adding an absolute minimum length of 2 for a motif. We continue training \foldflowpp for 330,000 steps on the same dataset, with a learning rate of 10e-5. This was done using 2 NVIDIA A100 80G, for 2.85 days.

\subsubsection{VHH CDR} \label{app:cdr-training}

From the Structural Antibody Database we used 615 nanobody sequences, yielding 1831 chains from the PDB as training set, and for testing we used 40 sequences appearing in 106 PDB chains. With both the sequence and CDRs readily available, we can build the appropriate mask for training, which is used to fix the motif atoms, and mask the scaffold sequence information. For testing, we sampled the scaffold segment lengths based on the median value of each scaffold segment, $\pm$ 5. Empirically the scaffold segment lengths varied less than that amount; nanobodies are known to exhibit less variability across their framework regions, both in sequence and structure \citep{mitchell2018comparative}. We continue training \foldflowpp using this dataset, training for 10,000 steps with a learning rate of 10e-5. Using a single A100 80G this process takes 3.5 hours. We observed rapid increase but rapid tapering in performance using the VHH dataset; this is likely due to the lack of variability of the scaffolds themselves.

\begin{figure}[ht]
    \centering
    \caption{Samples of scaffolds for VHHs. Motif (i.e. CDR) is in red, scaffold is in blue, and refolded structure from ESMFold is in gray.}
    \label{fig:vhh_samples}
    \begin{tabular}{c c}
        \includegraphics[height=4cm]{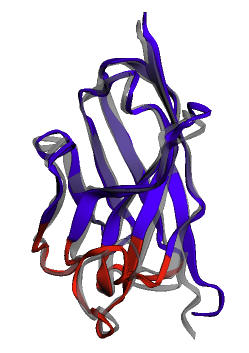} & 
        \includegraphics[height=4cm]{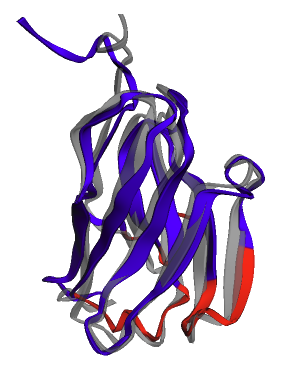} \\
        \includegraphics[height=4cm]{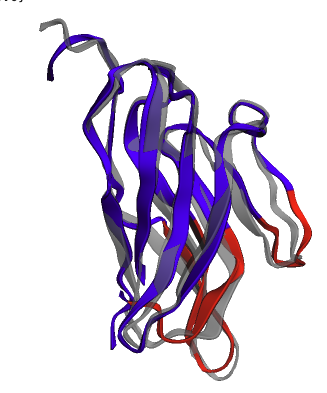} &
        \includegraphics[height=4cm]{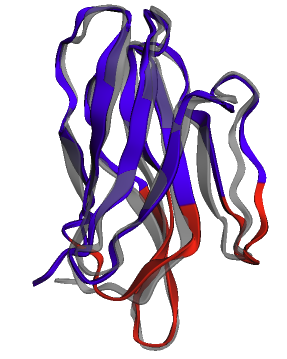}
    \end{tabular}
\end{figure}

We note the designability metric used here and in other papers shows certain limitations when applied to this dataset and this task. Using the test set sequence and structures themselves, we compute the same scRMSD scores, in order to benchmark what should be, in theory, the best possible performance. We observe that out of the 106 testing chains, only 25 of them are "solved" according to \cite{watson_novo_2023}'s criteria. This raises the question as to whether or not this particular criteria and setup is applicable to any motif scaffolding task. We provide the full set of results in \cref{tab:cdr_full} on all 106 chains, and on a subset of size 25 in \cref{tab:appendix_des_cdr_designable}, with generated samples in \cref{fig:vhh_samples}. In addition, the particular number of solved samples are reported in \cref{tab:motif_vhh_full}.

\begin{table}[ht]
    \centering
        \footnotesize
        \caption{\small VHH Motif Scaffolding results on the re-foldable examples only. Reported are the global, motif, and scaffold RMSD along with the number of solved tasks.}
        \label{tab:appendix_des_cdr_designable}
        \begin{tabular}{l C{2cm} C{2cm} C{2cm} C{2cm}}
        \toprule
         & Global scRMSD & Motif scRMSD & Scaffold scRMSD & Solved\newline(out of 25) \\
        \midrule
        RFDiffusion & 3.1$\pm$1.23 & 3.94$\pm$1.54 & 2.4$\pm$0.93 & 5 \\
        \foldflowpp VHH & 2.27$\pm$0.5 & 2.78$\pm$1.01 & 1.67$\pm$0.24 & 9 \\
        \midrule
        Test Set & 0.55$\pm$0.19 & 0.65$\pm$0.18 & 0.48$\pm$0.19 & 25 \\
        \bottomrule
        \end{tabular}
\end{table}

\begin{table}[ht]
    \centering
        \footnotesize
        \caption{\small VHH Motif Scaffolding results on all samples, same numbers as \cref{tab:appendix_des_cdr_designable}}
        \label{tab:cdr_full}
        \begin{tabular}{l C{2cm} C{2cm} C{2cm} C{2cm}}
        \toprule
         & Global scRMSD & Motif scRMSD & Scaffold scRMSD & Solved (out of 106) \\
        \midrule
        RFDiffusion & 2.86$\pm$1.1 & 3.6$\pm$1.42 & 2.25$\pm$0.85 & 18 \\
        \foldflowpp VHH & 2.3$\pm$0.47 & 2.99$\pm$0.86 & 1.66$\pm$0.31 & 15 \\
        \midrule
        Test Set & 1.49$\pm$1.38 & 2.17$\pm$1.22 & 1.05$\pm$1.69 & 25 \\
        \bottomrule
        \end{tabular}
\end{table}

\subsection{Zero-shot Molecular Dynamics}
\label{app:conformation_sampling}
To evaluate \foldflowpp's ability to capture protein dynamics we evaluated its performance on the test set of ATLAS \citep{10.1093/nar/gkad1084} molecular dynamics used by \cite{jing2024alphafold} but restricted to proteins at most 40 amino acids in length. To measure performance we used the following metrics. All metrics were computed exclusively over backbone atoms.

\begin{enumerate}
    \item \textbf{Pairwise RMSD $r$.} To measure \foldflowpp's ability to capture protein flexibility we first compute the average pairwise RMSD between every pair of conformations generated by each method. We then evaluate the average pairwise RMSD for the ground truth data and report the Pearson correlation $r$ between the average pairwise RMSD per generated protein and ground truth data.
    \item \textbf{Global and per-target RMSF $r$.} To further investigate flexibility, we measure the RMSF both globally and per target and compute the Pearson correlation $r$ to ground truth data. Global RMSF is computed by, for each target calculating the backbone RMSF and taking its average, then measuring Pearson correlation between generated and ground truth samples over the sequence of averaged RMSFs. For per-target RMSF we instead compute the Pearson correlation between generated and ground truth backbone atom RMSFs and report the average taken over all targets.
    \item \textbf{PCA $\mathcal{W}_2$.} Following \cite{jing2024alphafold} we seek to measure the distributional accuracy of our generated samples by evaluating the 2-Wasserstein distance using the first two principal components given by ground truth data. We use a PCA as evaluating $\mathcal{W}_2$ on all atom coordinates would yield inaccurate measurements due to $\mathcal{W}_2$ needing samples exponential in dimensionality in order to obtain reasonable estimates. We compute the PCA $\mathcal{W}_2$ by, for each target, computing the first two principal components of backbone atom coordinates from ground truth MD data and projecting the backbone coordinates of each method's generated samples onto these coordinates.  We then compute the $\mathcal{W}_2$ per target and report the averaged $\mathcal{W}_2$ over all targets.
\end{enumerate}

As the test set contains 30,000 frames per protein computing test metrics using all ground truth conformations would be computationally infeasible.  As such, following \cite{jing2024alphafold} we randomly sample 300 conformations for each protein to be used as the test set. \cref{tab: md-task} reports the mean and standard deviation over 5 resamplings of these test sets. Samples were generated from \foldflowpp using 50 inference steps and the inference annealing trick wherein the rotation vector is multiplied by $10t$.

\begin{figure}
    \label{fig:more_md_samples} 
    \vspace{-15pt}
        \centering
        \includegraphics[width=1.0\textwidth]{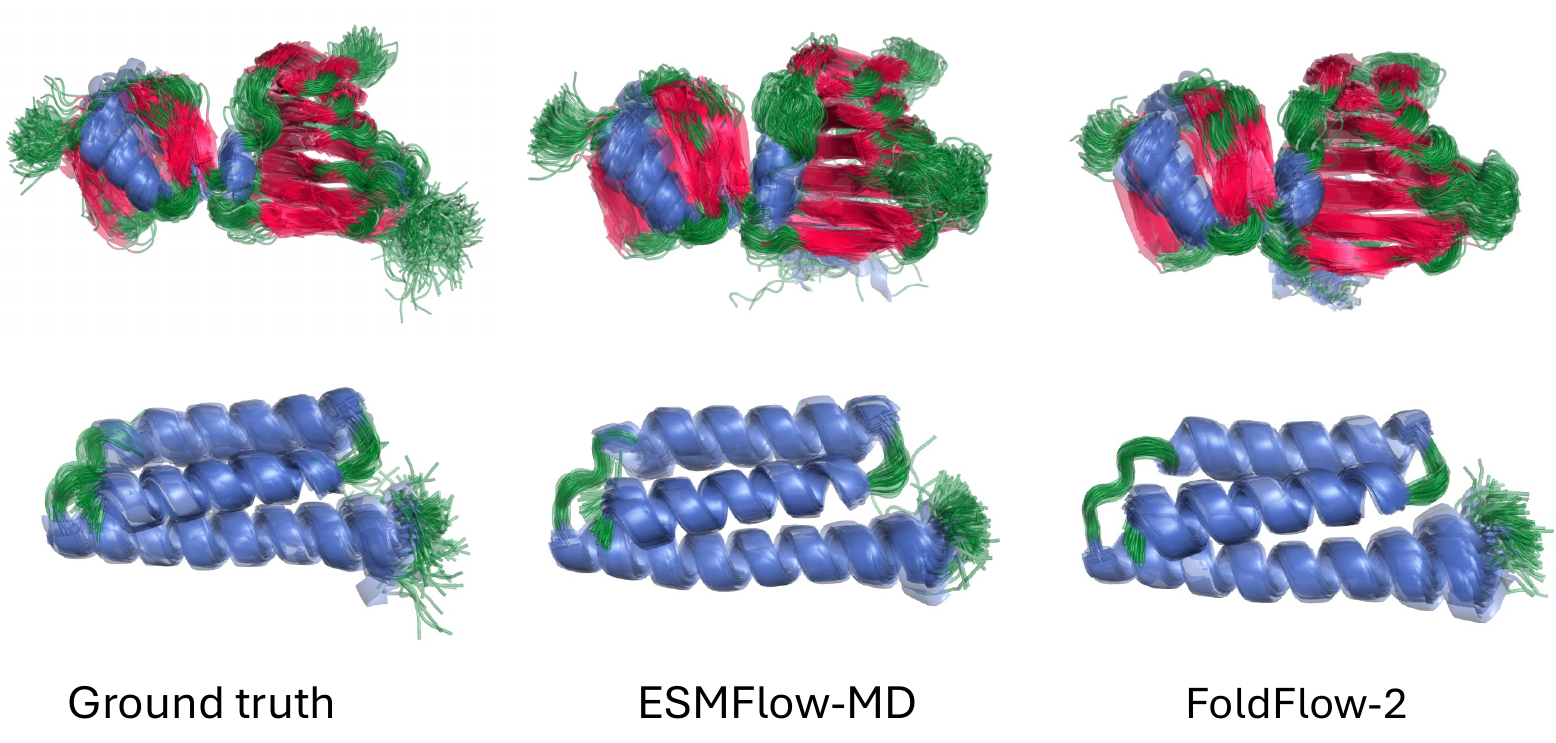}
        \caption{\small Additional conformation generation task samples. Proteins are colored by secondary structure with $\alpha$-helices in blue, $\beta$-sheets in red, and loops in green.}
        \vspace{-5pt}
\end{figure}

We report in \cref{tab:md-speed} details on the resources required for training \foldflowpp compared to AlphaFlow-MD and ESMFlow-MD. We see that \foldflowpp requires an order of magnitude less time per inference step than ESMFlow-MD and AlphaFlow-MD while attaining results competitive with ESMFlow-MD while using 4.5X less GPU hours for training and 33X less trainable parameters. \foldflowpp, ESMFlow-MD, and AlphaFlow-MD were all done on NVIDIA A100s.  Inference time benchmarks were done on an NVIDIA A100, performing inference on a single protein of length 300 amino acids. Finally, in \cref{fig:more_md_samples} we provide additional generated conformational ensembles from the test set, ESMFlow, and \foldflowpp.

\begin{table}[ht!]
    \vspace{-10pt}
    \caption{\small Molecular dynamics experiment training details.}\label{tab:md-speed}
    \centering
    \resizebox{\textwidth}{!}{
    \begin{tabular}{lrrrr}
        \toprule
        Model & \# training GPU hours & \# total parameters & \# trainable parameters & Inference time / step (sec) \\
        \midrule
        AlphaFlow-MD  & 2224 & 95M & 95M  & 3.26 $\pm$ 0.01\\
        ESMFlow-MD  & 872 & 3.5B & 694M  & 1.12 $\pm$ 0.01\\
        \foldflowpp  & 192 & 672M & 21M & 0.18 $\pm$ 0.00\\
        \bottomrule
    \end{tabular}
    }
\end{table}

\section{Additional Results}\label{app:additional_results}

\begin{figure}[ht]
    \centering
    \includegraphics[width=0.8\textwidth]{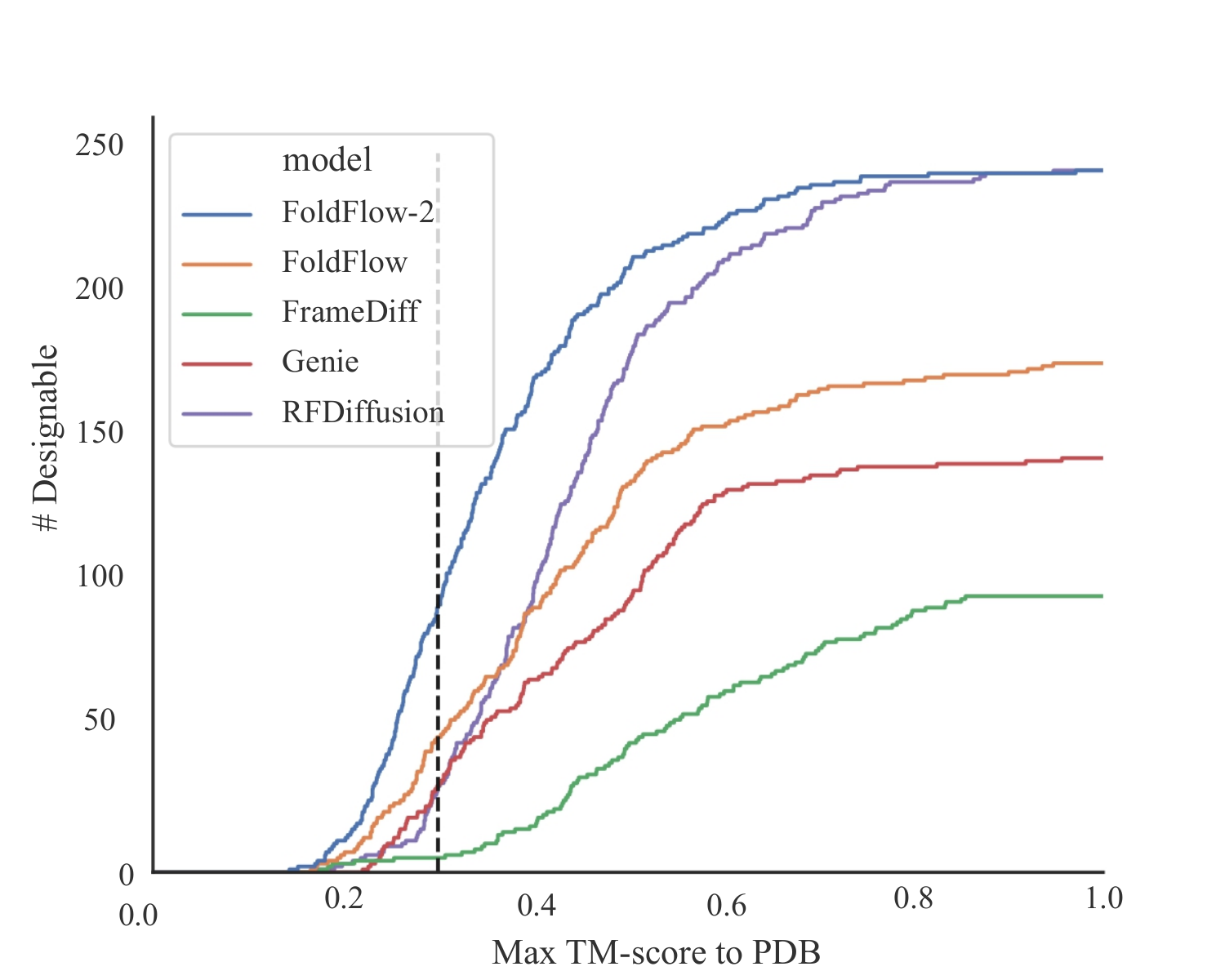}
    \caption{Curve showing the number of designable proteins that are at least some distance away from the PDB. \foldflowpp has many more novel and designable proteins than baselines. We report designability fraction at TM-score = 0.3 in \cref{tab:main_table}.}
    \label{fig:designability_novelty}
\end{figure}

\subsection{Unconditional generation diversity and novelty exploration}

\begin{figure}[ht]
    \centering
    \includegraphics[width=1\textwidth]{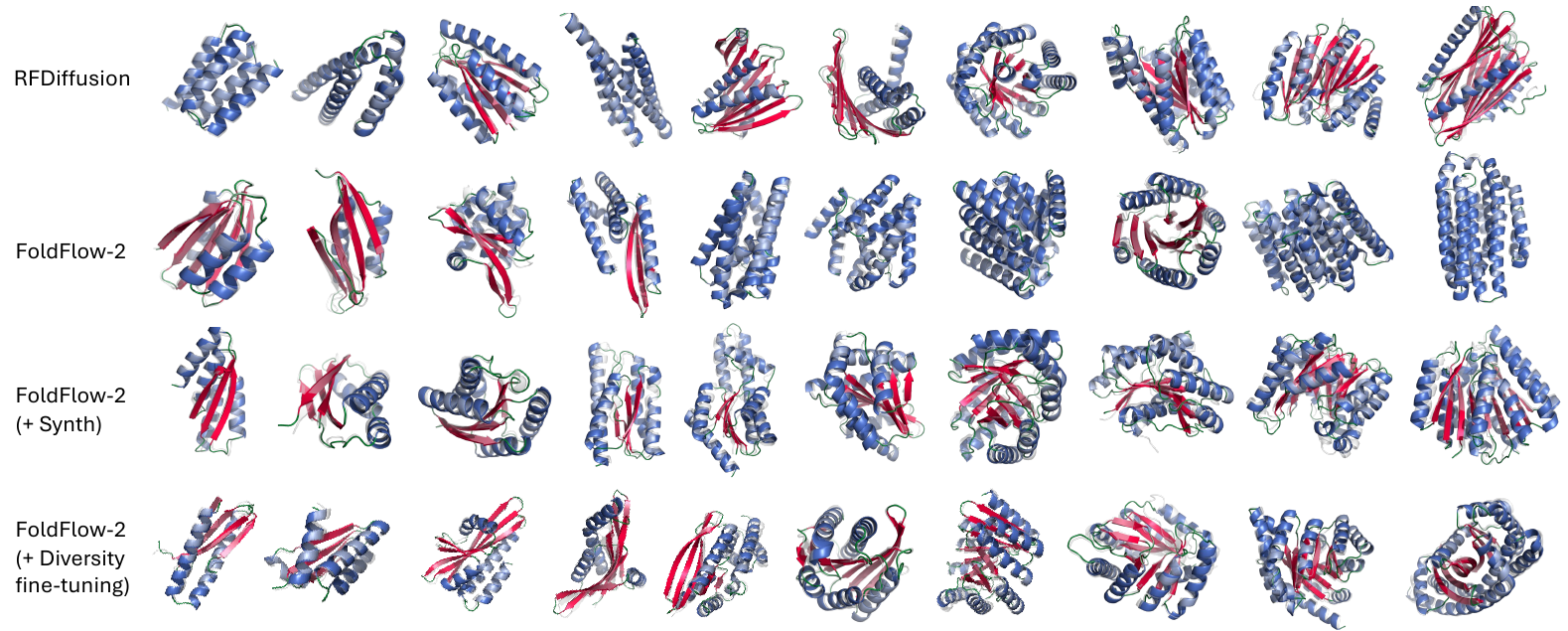}
    \caption{Designable samples from various methods. Overlayed in silver are refolded ESMFold structures. \foldflowpp exhibits significantly more diversity in secondary structure at shorter lengths than RFDiffusion with fine-tuned models able to produce diverse proteins across lengths.}
    \label{fig:generated_samples_ff}
\end{figure}

We next provide several additional results on unconditional generation to give a better understanding of the behavior of \foldflowpp relative to the baselines. In \cref{fig:designability_novelty} we can see that \foldflowpp creates many more novel proteins at \textit{all thresholds} of what is considered novel as compared to previous methods. We also depict more generated samples of all lengths in \cref{fig:generated_samples_ff}. We can see that \foldflowpp creates more diversity of structures, especially at shorter lengths. With synthetic data or diversity fine-tuning this is expanded to all lengths 100-300. 

\subsection{Ablation study and sensitivity analysis}\label{sec:ablation}

In this section, we provide several ablations of our \foldflowpp method. In \cref{tab:ablation}, we provide the unconditional generation performance for the different architecture components of our \foldflowpp method. In \cref{tab:annealing_inference_t} and \cref{tab:app-folding-rmsd-speed}, we compare the performance achieved by our model \foldflowpp when we use different inference annealing $t$ values for respectively unconditional generation and folding. In \cref{tab:euler_steps_unconditional_gen} and in \cref{tab:app-folding-rmsd-steps}, we compared the performance achieved by our model \foldflowpp when using different numbers of Euler steps at inference for respectively unconditional generation and folding.

\xhdr{Architecture ablation} In \cref{tab:ablation} we seek to understand the effects of architecture and dataset on the performance across our main designability, novelty, and diversity metrics for unconditional backbone generation. Starting from \foldflowpp we first investigate the effect of replacing the Folding Block with a simple MLP with \foldflowpp (- F.\ Block) and removing the sequence conditioning entirely (- ESM2). In this comparison we find that both the folding block and structure conditioning significantly improve the results. We find that the Folding block improves all metrics while the structure conditioning improves designability and diversity at the cost of novelty.

\xhdr{Dataset and Training} In \cref{tab:ablation}, we look at how adding synthetic data or stochastic flow matching affects designability, diversity, and novelty metrics. We find that both these additions actually hurt these metrics, although this is likely due to the change in composition of their generated structures. Overall, we find that these two models create more diverse proteins.

\begin{figure}[ht]
    \centering
    \begin{subfigure}[b]{0.245\textwidth}
        \centering
        \includegraphics[width=1\textwidth]{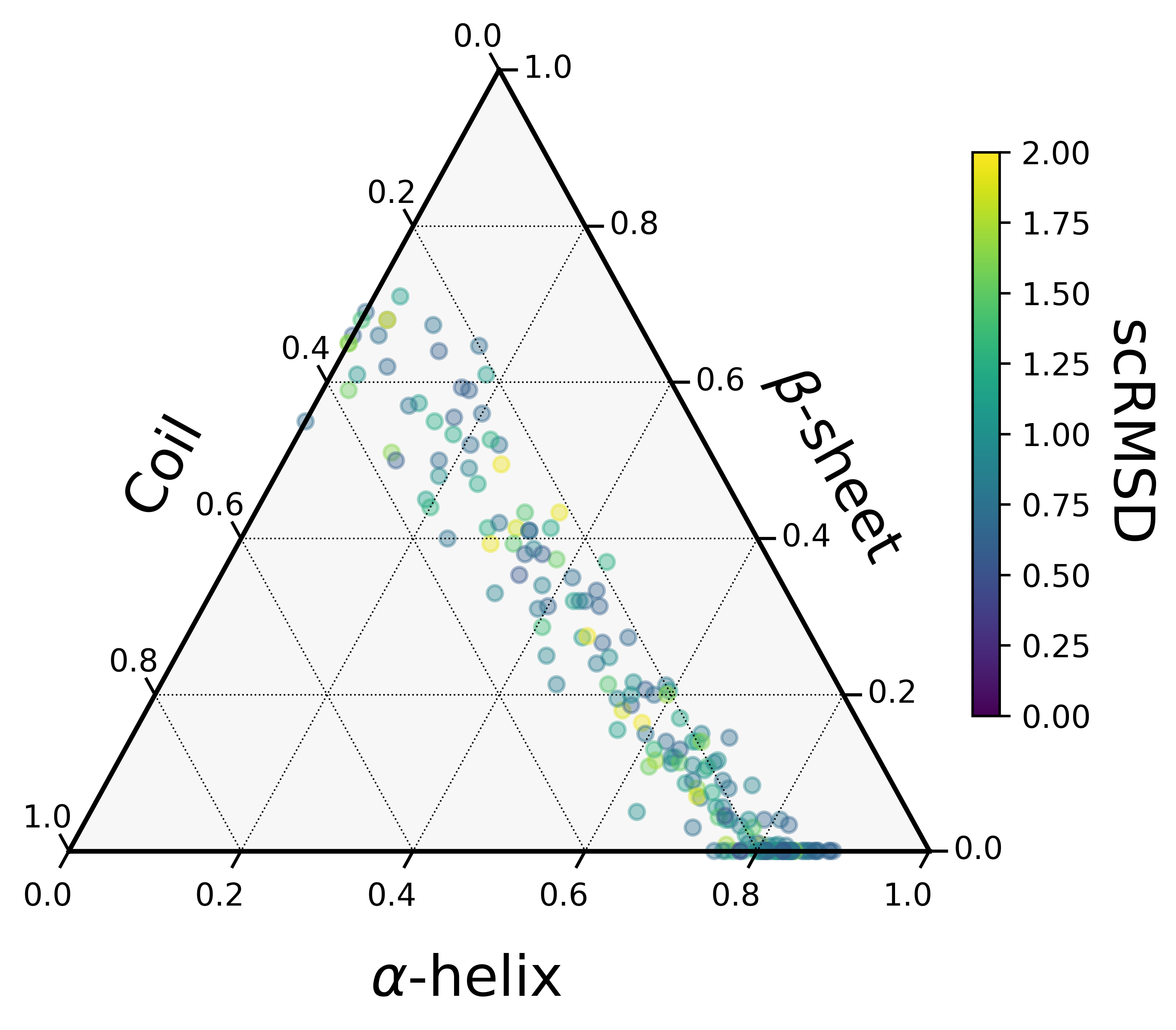}
        \caption{100 steps}
    \end{subfigure}
    \begin{subfigure}[b]{0.245\textwidth}
        \centering
        \includegraphics[width=1\textwidth]{figures/filtered/ffpp/secondary_structure_plot.png}
        \caption{50 steps}
    \end{subfigure}
    \begin{subfigure}[b]{0.245\textwidth}
        \centering
        \includegraphics[width=1\textwidth]{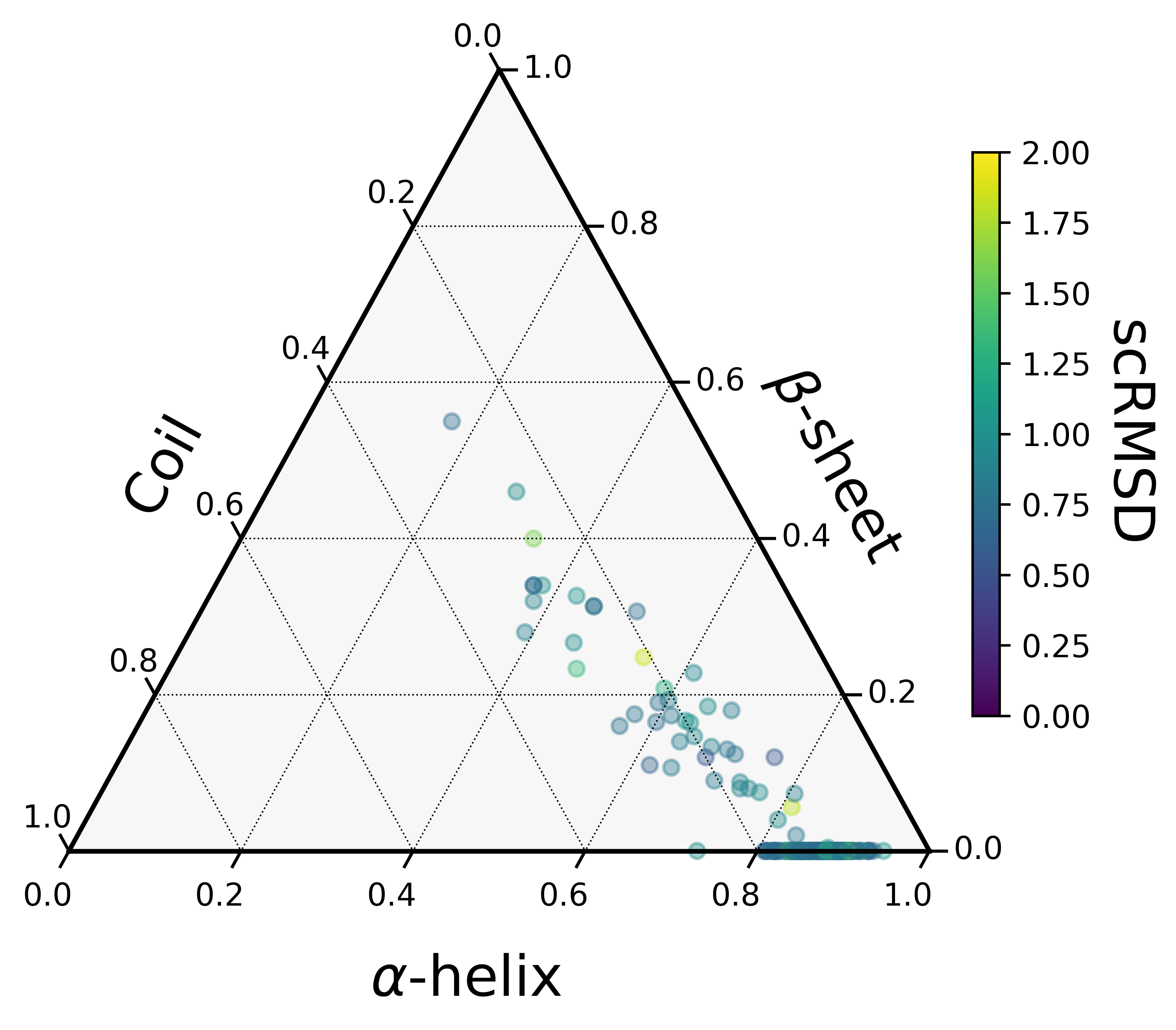}
        \caption{30 steps}
    \end{subfigure}
    \begin{subfigure}[b]{0.245\textwidth}
        \centering
        \includegraphics[width=1\textwidth]{figures/training_triangle.png}
        \caption{Data}
    \end{subfigure}
    \vspace{-10pt}
    \caption{\small Secondary structure elements distributions ($\alpha$-helices, $\beta$-sheets, and coils) of designable (scRMSD $< 2.0$) proteins, along with the data's distribution. We find more steps leads to more diverse designs with fewer $\alpha$-helix only generations.}
        \label{fig:secondary_structure_diversity}
    \vspace{-15pt}
\end{figure}

\xhdr{Number of inference steps \& inference annealing scale}
We also studied the influence of number of Euler steps on the generated proteins in \cref{tab:euler_steps_unconditional_gen}. We note that our model performs quite well at a relatively small number of steps, although performance starts to drop off under 25 steps. We attribute this to the optimal transport approach which is known to increase quality of generation especially with few inference steps~\cite{tong2023improving,pooladian_2023_multisample}. We find an interesting tradeoff between designability and diversity in \cref{tab:euler_steps_unconditional_gen} and visually in \cref{fig:secondary_structure_diversity}. Specifically that more steps increases the diversity of samples at the cost of designability.

We also studied the impact of the inference annealing scale factor in \cref{tab:annealing_inference_t}. We see that small scaling (or none at all, corresponding to a value of $1$) produce highly undesignable proteins, but designability improves quickly and beyond the optimal value of $10$ it is somewhat stable. We notice the opposite effect in diversity as measured by the MaxCluster metric: no time scaling yields a score which is 64\% larger than the same metric with scaling $10$, and this trend is clearly anti-correlated with the scaling value.

\xhdr{Impact of synthetic augmented dataset on diversity}
One of the interesting benefits of using our synthetic augmented dataset is that it increases diversity among designable generated data. Proteins have different secondary structures such as helices, beta-sheets and coil. While our model generates a lot of helices, it is important for drug discovery applications to generate other secondary structure such as beta-sheets. As shown in \cref{fig:secondary_structure_diversity}, we can see that our synthetic augmented dataset leads to an increased of beta-sheets. 

\xhdr{Impact of rotation time scaling \& number of inference steps on folding}
We conducted an analysis of inference parameters on \foldflowpp's ability to fold proteins. In \cref{tab:app-folding-rmsd-speed} we sweep over the rotation time scaling parameter and measure its impact on folding RMSD. We see a similar trend to the unconditional case in \cref{tab:annealing_inference_t}: very small scaling factors (e.g., 2) produce worse results, but the performance improves quickly as scaling increases and remains somewhat stable in a neighbourhood of 10.

We also conducted a sweep over the number of inference steps during generation on folding RMSD in \cref{tab:app-folding-rmsd-steps}. We again see a similar trend to the counterpart for unconditional generation in \cref{tab:euler_steps_unconditional_gen}: a small number of Euler steps near 30-50 is best, with performance degrading significantly beyond 50 steps.

\begin{table}[ht!]
    \centering
    \caption{Ablation study on \foldflowpp (FF-2) using: synthetic data, folding blocks, and stochastic flow matching (SFM). We generated 250 proteins (50 of length 100, 150, 200, 250) and compared Designability (fraction with scRMSD < $2.0\angstrom$), Novelty (max. TM-score to PDB and fraction of proteins with averaged max.\ TMscore $<0.3$ and scRMSD < $2.0\angstrom$), and Diversity (avg. pairwise TMscore and MaxCluster fraction).}
    \resizebox{\textwidth}{!}{
    \begin{tabular}{l c c c c c | c c c |c c}
        \toprule
        & \multicolumn{1}{c}{Designability}  & \multicolumn{2}{c}{Novelty} & \multicolumn{2}{c}{Diversity} & Seq. & Folding & SFM & iter/s & \# train\\
        & Frac.\ $<2\angstrom$ ($\uparrow$)  & Frac.\ TM $< 0.3$ ($\uparrow$) & avg.\ max TM ($\downarrow$) & p.wise TM ($\downarrow$)& MaxClust. ($\uparrow$) &  cond. & blocks & & & param.\\
        \midrule
        FF-2   (- F.\ Block - ESM2) & 0.716 $\pm$ 0.029 & 0.188 $\pm$ 0.025 & 0.419 $\pm$ 0.012 & 0.240 & 0.228 &\xmark & \xmark & \xmark & 2.7 & 17M \\
	FF-2 (- F.\ Block) & 0.852 $\pm$ 0.023 & 0.148 $\pm$ 0.023 & 0.438 $\pm$ 0.010 & 0.227 & 0.271 &\cmark & \xmark & \xmark & 2.1 & 18M\\
        FF-2 & \textbf{0.976 $\pm$ 0.010} & \textbf{0.368 $\pm$ 0.031} & \textbf{0.363 $\pm$ 0.009} & \textbf{0.205} & \textbf{0.348} &\cmark & \cmark & \xmark & 1.6 & 21M \\
        FF-2 (+Synthetic) & 0.785 $\pm$ 0.027 & 0.047 $\pm$ 0.014 & 0.465 $\pm$ 0.008 & 0.226 & 0.264 &\cmark & \cmark & \xmark & 1.6 & 21M \\
        FF-2 (+SFM) & 0.935 $\pm$ 0.016 & 0.274 $\pm$ 0.029 & 0.386 $\pm$ 0.009 & 0.218 & 0.281 &\cmark & \cmark & \cmark & 1.5 & 21M\\
        \bottomrule
    \end{tabular}
    }
    \label{tab:ablation}
\end{table}

\begin{table}[htb]
    \centering
    \vspace{-15pt}
    \caption{\looseness=-1\small Comparison of Designability (fraction with scRMSD < $2.0\angstrom$), Novelty (max. TM-score to PDB and fraction of proteins with averaged max.\ TMscore $<0.3$ and scRMSD < $2.0\angstrom$), and Diversity (avg. pairwise TMscore and MaxCluster fraction) for different inference annealing functions $i(t)$.}
    \resizebox{1\textwidth}{!}{
    \begin{tabular}{lcccccc}
        \toprule
          \multicolumn{1}{c}{} & \multicolumn{1}{c}{Designability}  & \multicolumn{2}{c}{Novelty} & \multicolumn{2}{c}{Diversity} \\
          \cmidrule(lr){2-2} \cmidrule(lr){3-4} \cmidrule(lr){5-6}
        & Frac.\ $<2\angstrom$ ($\uparrow$)  & Frac.\ TM $< 0.3$ ($\uparrow$) & avg.\ max TM ($\downarrow$) & pairwise TM ($\downarrow$)& MaxCluster ($\uparrow$)\\
        \midrule
$1$ & 0.104 $\pm$ 0.019 & 0.012 $\pm$ 0.007 & 0.427 $\pm$ 0.022 & 0.197& \textbf{0.571}\\
$2t$ & 0.148 $\pm$ 0.023 & 0.040 $\pm$ 0.012 & 0.403 $\pm$ 0.024 & 0.198& 0.549\\
$5t$ & 0.832 $\pm$ 0.024 & 0.312 $\pm$ 0.029 & \textbf{0.375 $\pm$ 0.011} & \textbf{0.193} & 0.387\\
$10t$ & \textbf{0.976 $\pm$ 0.010} & \textbf{0.368 $\pm$ 0.031} & 0.363 $\pm$ 0.009 & 0.205 & 0.348\\
$15t$ & 0.928 $\pm$ 0.016 & 0.308 $\pm$ 0.029 & 0.388 $\pm$ 0.010 & 0.199& 0.358\\
$20t$ & 0.944 $\pm$ 0.015 & 0.304 $\pm$ 0.029 & 0.395 $\pm$ 0.010 & 0.199& 0.347\\
        \bottomrule
        \end{tabular}
        }
    \label{tab:annealing_inference_t}
    \vspace{-10pt}
\end{table}

\begin{table}[]
    \centering
    \caption{\small Speed of the integration on rotations. Integrating with a faster time for rotations compared to translation leads to more designable structures. Reporting the mean $\pm$ std. on 278 test samples.}
    \label{tab:app-folding-rmsd-speed}
\begin{tabular}{ll}
\toprule
 Rotation time scaling & RMSD ($\downarrow$) \\
\midrule
$2t$ & 3.641 $\pm$ 4.457 \\
$5t$ & 3.257 $\pm$ 4.113 \\
$10t$ & 3.334 $\pm$ 4.325 \\
$15t$ & \textbf{3.237} $\pm$ \textbf{4.145} \\
\bottomrule
\end{tabular}
\end{table}

\begin{table}[htb]
    \centering
    \vspace{-15pt}
    \caption{\looseness=-1\small Comparison of Designability (fraction with scRMSD < $2.0\angstrom$), Novelty (max. TM-score to PDB and fraction of proteins with averaged max.\ TMscore $<0.3$ and scRMSD < $2.0\angstrom$), and Diversity (avg. pairwise TMscore and MaxCluster fraction) for different number of Euler steps at inference.}
    \resizebox{1.0\textwidth}{!}{
    \begin{tabular}{lcccccc}
        \toprule
          \multicolumn{1}{c}{} & \multicolumn{1}{c}{Designability}  & \multicolumn{2}{c}{Novelty} & \multicolumn{2}{c}{Diversity} \\
          \cmidrule(lr){2-2} \cmidrule(lr){3-4} \cmidrule(lr){5-6}
        & Frac.\ $<2\angstrom$ ($\uparrow$)  & Frac.\ TM $< 0.3$ ($\uparrow$) & avg.\ max TM ($\downarrow$) & pairwise TM ($\downarrow$)& MaxCluster ($\uparrow$)\\
        \midrule
15 Euler steps & 0.480 $\pm$ 0.032 & 0.136 $\pm$ 0.022 & 0.382 $\pm$ 0.012 & 0.196 & 0.430\\
20 Euler steps &	0.876 $\pm$ 0.021 & 0.328 $\pm$ 0.030 & 0.358 $\pm$ 0.009 & 0.203 & 0.341\\
25 Euler steps & 0.948 $\pm$ 0.014 & 0.376 $\pm$ 0.031 & 0.372 $\pm$ 0.010 & 0.207 & 0.336\\
30 Euler steps & 0.976 $\pm$ 0.010 & 0.368 $\pm$ 0.031 & 0.363 $\pm$ 0.009 & 0.205 & 0.348\\
35 Euler steps & \textbf{0.980} $\pm$ 0.009& 0.356 $\pm$ 0.030 & 0.370 $\pm$ 0.008 & 0.210 & 0.305\\
40 Euler steps &0.960 $\pm$ 0.012 & 0.304 $\pm$ 0.029 & 0.382 $\pm$ 0.009 & 0.201 & 0.349\\
45 Euler steps & 0.940 $\pm$ 0.015 & 0.328 $\pm$ 0.030 & 0.382 $\pm$ 0.009 &0.201 & 0.357 \\
50 Euler steps & 0.952 $\pm$ 0.014 & \textbf{0.384 $\pm$ 0.031} & 0.383 $\pm$ 0.011 & 0.194 & 0.387\\
75 Euler steps & 0.800 $\pm$ 0.025 & 0.300 $\pm$ 0.029 & 0.380 $\pm$ 0.010 & 0.189 & 0.434\\
100 Euler steps & 0.748 $\pm$ 0.028 & 0.340 $\pm$ 0.030 & \textbf{0.366 $\pm$ 0.011} & 0.188 & 0.415\\
150 Euler steps & 0.620 $\pm$ 0.031 & 0.180 $\pm$ 0.024 & 0.408 $\pm$ 0.013 & 0.185 & 0.437\\
200 Euler steps & 0.580 $\pm$ 0.031 & 0.208 $\pm$ 0.026 & 0.391 $\pm$ 0.012 & \textbf{0.183} & \textbf{0.447}\\
        \bottomrule
        \end{tabular}
        }
    \label{tab:euler_steps_unconditional_gen}
    \vspace{-10pt}
\end{table}

\begin{table}[]
    \centering
    \caption{\small Effect of the number of integration steps on the aligned RMSD between the generated and ground truth backbone. Reporting the mean $\pm$ std.\ on 278 test samples.}
    \label{tab:app-folding-rmsd-steps}
\begin{tabular}{ll}
\toprule
 \# Euler steps & RMSD ($\downarrow$) \\
\midrule
30 Euler steps & 3.384 $\pm$ 4.223 \\
50 Euler steps & \textbf{3.334 $\pm$ 4.325} \\
100 Euler steps & 3.374 $\pm$ 4.377 \\
150 Euler steps & 3.405 $\pm$ 4.409 \\
200 Euler steps & 3.481 $\pm$ 4.465 \\
\bottomrule
\end{tabular}
\end{table}

\end{document}